  \RequirePackage{fix-cm}
  \documentclass[twocolumn]{svjour3}          
  \smartqed  
  \usepackage{graphicx}
  \usepackage[authoryear]{natbib}
  \usepackage{multicol}
  \usepackage{tabu,multirow}
  \usepackage[leftcaption]{sidecap}
  \usepackage{amsmath}
  \usepackage{amssymb}
  \usepackage{color}
  \usepackage{enumitem}
  \usepackage{verbatim}
  \usepackage[colorlinks,citecolor=blue]{hyperref}
  \usepackage{soul}
  \usepackage{graphicx}
  \usepackage{subfigure}
  \renewcommand{\arraystretch}{1.3} 
  \usepackage{float}  
  \usepackage{textcomp}
  
  \usepackage{multicol}
  \usepackage{booktabs}

  \usepackage{eso-pic}
  \usepackage{xspace}
  \makeatletter
  \DeclareRobustCommand\onedot{\futurelet\@let@token\@onedot}
  \def\@onedot{\ifx\@let@token.\else.\null\fi\xspace}

  \makeatother
  
  \usepackage{mathtools}

  \soulregister{\citep}7 

  \soulregister{\ref}7 

\begin{document}
  \sloppy
  
  \title{Intriguing Property  and Counterfactual Explanation of GAN for Remote Sensing Image Generation
  }
  
  
  
  \author{ Xingzhe Su      \and
           Wenwen Qiang \and
           Jie Hu \and
           Changwen Zheng \and
           Fengge~Wu \and
           Fuchun Sun
  }
  
 \institute{Xingzhe Su, Wenwen Qiang, Changwen Zheng, Fengge Wu \at
              Science \& Technology on Integrated Information System Laboratory, Institute of Software Chinese Academy of Sciences, University of Chinese Academy of Sciences, Beijing, China\\
           \and
           Jie Hu \at
           Meituan, Beijing, China\\
           \and
           Fuchun Sun \at
              Science \& Technology on Integrated Information System Laboratory, Department of Computer Science and Technology, Tsinghua University, Beijing, China\\
          \and
          Corresponding author: Wenwen Qiang, \email{qiangwenwen@iscas.ac.cn}
}

\date{Received: date / Accepted: date}

\def\ourconv{RIConv++\xspace}
\def\smallgap{\vspace{0.05in}}

  \maketitle
  
  \begin{abstract}
  
  Generative adversarial networks (GANs) have achieved remarkable progress in the natural image field. However, when applying GANs in the remote sensing (RS) image generation task, an extraordinary phenomenon is observed: the GAN model is more sensitive to the amount of training data for RS image generation than for natural image generation (Fig.\ref{fig1}). In other words, the generation quality of RS images will change significantly with the number of training categories or samples per category. In this paper, we first analyze this phenomenon from two kinds of toy experiments and conclude that the amount of feature information contained in the GAN model decreases with reduced training data (Fig.\ref{fig2}). Then we establish a structural causal model (SCM) of the data generation process and interpret the generated data as the counterfactuals. Based on this SCM, we theoretically prove that the quality of generated images is positively correlated with the amount of feature information. This provides insights for enriching the feature information learned by the GAN model during training. Consequently, we propose two innovative adjustment schemes, namely Uniformity Regularization (UR) and Entropy Regularization (ER), to increase the information learned by the GAN model at the distributional and sample levels, respectively. Extensive experiments on eight RS datasets and three natural datasets show the effectiveness and versatility of our methods. The source code is available at \href{https://github.com/rootSue/Causal-RSGAN}{https://github.com/rootSue/Causal-RSGAN}.
  
  \end{abstract}




\section{Introduction}
\label{sec:intro}

The image generation task \citep{xu1996convergence,hinton2006reducing,ranzato2010generating} aims to learn the real distribution of images, builds effective generation models, and produces various realistic images by changing some potential parameters. Generative adversarial networks (GANs) \citep{10.5555/2969033.2969125} have made significant progress as an efficient solution to this task, especially in synthesizing high-fidelity images. Nowadays, the GAN models have become the cornerstone techniques for numerous vision applications, such as image super-resolution \citep{srgan,blind,he2022gcfsr}, image inpainting \citep{ip0,ip1}, domain adaptation \citep{coGAN, da1}, and image-to-image translation \citep{GauGAN,SketchGAN}.


GAN is also a favored method for the generation task in the remote sensing (RS) field. The applications of GANs in RS include augmenting training samples in classification \citep{MartaGAN, AttentionGAN, MFGAN}, change detection \citep{chen2021adversarial}, and dealing with image-generation-related tasks such as image translation \citep{bejiga2020improving, disaster, smapgan,deepfake2021} and image super-resolution \citep{jiang2019edge,xiong2020improved}. However, the RS images synthesized by these methods are of low quality. To tackle this issue, an alternative is to directly apply the well-established GAN models to RS image generation tasks. During the experiments, we found an unusual phenomenon that has not been seen in natural image generation tasks, that is, the generation quality of remote sensing images will change significantly with different amount of training data.

To be more specific, we perform image generation experiments on the NWPU-RESISC45 (NWPU) \citep{NWPU} and PatternNet (PN) \citep{patternnet} datasets in the same settings as previous work \citep{transitionCGAN}. We gradually reduce the training data size by reducing the number of classes or the number of samples per class. In both cases, the Frechet Inception Distance score (FID) \citep{FID} gradually increases, indicating worse generation quality. The overall experiment results in Fig.\ref{fig1} are consistent with those of natural images. However, we find a unique phenomenon: the GAN model is more sensitive to the size of training data for RS image generation than for natural image generation. The FID varies more significantly than those of natural images under different numbers of training classes or samples per class.

To find the reasons behind this, we analyze the features of these generated samples (Fig.\ref{fig2}). We observe that the features get sparser as the number of training classes decreases, and the distributions of generated samples in the feature space get uneven when reducing training samples per class. Inspired by the information theory, sparser features contain less information, and the uniform distribution gets the maximum information entropy \citep{information}.
Therefore, for the sensitivity of the GAN model to the training data size, a plausible interpretation is that the amount of the feature information contained in the GAN model dramatically decreases when training data sizes are reduced, leading to poor generation quality. To further interpret the relation between the feature information and the generation quality, we establish a structural causal model (SCM) of data generation process. We interpret the generated images as the counterfactuals and theoretically prove that the quality of generated images is positively correlated with the amount of feature information. This provides insights to improve the generation quality of RS images by enriching the amount of feature information contained in the GAN model during training. 

In this paper, we propose distribution-wise and sample-wise adjustment schemes to increase the information entropy so as to enrich the amount of feature information contained in the GAN model during training. Specifically, based on our discovery (Fig.\ref{fig2}), we impose a uniformity regularization on the distribution of features of generated samples and an entropy regularization on the features of the individual sample. We empirically show that these two regularization terms are effective not only in remote sensing image generation tasks, but also on natural images. 
In summary, the contributions of this paper are the following:
\begin{itemize}
\item We explore the properties of GAN models on RS image generation tasks and find that the GAN model is more sensitive to the amount of training data for RS image generation than for natural image generation.
\item After experiments, we conclude that the amount of feature information contained in the GAN model decreases dramatically with reduced training dataset, leading to poor generation quality.
\item We establish a structural causal model (SCM) of the data generation process and interpret the generated images as the counterfactuals. Based on this SCM, we theoretically prove that the quality of generated images is positively correlated with the amount of feature information.
\item Based on the counterfactual theory, we propose distribution-wise and sample-wise regularization schemes, which can be applied to arbitrary GAN models, to increase the feature information learned by the models.
\item The effectiveness and versatility of our methods have been demonstrated through extensive experiments on various GAN approaches, as well as remote sensing and natural datasets.

\end{itemize}

\section{Related Work}
\textbf{Generative adversarial networks}. Based on the idea of the zero-sum game, GANs aim to model the target distribution using adversarial learning. Various modifications have been proposed to stabilize the training process and improve the quality of the generated samples, including training objectives \citep{WGAN,WGAN-GP,LSGAN,SNGAN,r1,RGAN} and network architectures \citep{SAGAN,biggan,StyleGAN,StyleGAN2,esser2021taming,jiang2021transgan}. Among these methods, BigGAN \citep{biggan} and StyleGAN2 \citep{StyleGAN2} stand out because of the high-quality synthetic samples and the broad applications. Besides, several studies \citep{gulrajani2018towards, webster2019detecting} raise the concern of insufficient data for training the GAN models. Recent research exploits data augmentation \citep{zhao2020differentiable,stylegan2-ada,APA} methods to increase data diversity. Besides, Liu et al. \citep{liu2020towards} proposed a lighter network architecture and a self-supervised discriminator. Tseng et al. proposed the LeCam \citep{lecam} regularization to prevent the discriminator from over-fitting on small data. Coupled with these methods, the performance of GAN models becomes more outstanding. However, current GAN models can generate high-resolution natural images but not realistic remote sensing images.

\textbf{GAN in the RS field}. GANs have been applied to various kinds of RS images, such as infrared images \citep{infrared}, hyperspectral images \citep{SpecGAN,multispectral} and synthetic aperture radar (SAR) images \citep{2019Integrated}. In this paper, we focus on RS RGB images. Existing GAN models in the RS field can be divided into two types according to their applications. This first kind is augmenting training samples \citep{MartaGAN,AttentionGAN,MFGAN, chen2021adversarial}. Lin et al. propose MARTAGAN \citep{MartaGAN}, which is the first time that GANs have been applied to RS images. Chen et al. propose IAug \citep{chen2021adversarial}, which leverages the GAN model to generate RS images that contain changes involving plenty and diverse buildings. The second kind deals with image-generation-related tasks such as image translation \citep{ bejiga2020improving,disaster,smapgan,deepfake2021} and image super-resolution \citep{jiang2019edge,xiong2020improved}. However, these methods do not delve into the characteristics of RS images, and the generated images are of low quality. Su et al. propose GSGAN \citep{gsgan} to generate controllable and realistic RS images. This is the first work that focuses on RS image generation tasks. However, its improvements mostly rely on intuition and lack theoretical support.

As far as we know, this paper is the first to explore the properties of GAN models in RS image generation tasks. We discover an overlooked phenomenon: GAN models exhibit greater sensitivity to the amount of training data in RS image generation compared to natural image generation (Section \ref{sec:pre}). We establish a SCM of the data generation process and conduct an analysis of this phenomenon using counterfactual theory (Section \ref{sec:causal}). Building upon this analysis, we introduce two innovative adjustment schemes (Section \ref{sec:methods}). We empirically prove our methods are effective, versatile, and require neither modification of the network structure nor excessive computational effort (Section \ref{sec:ex})).

\section{Problem Study and Analysis}
\label{sec:pre}

In this section, we first introduce the basic framework of GAN. Then we explore the properties of GAN models on RS image generation tasks and find an unusual phenomenon that has not been observed on natural images. Finally, we analyze this phenomenon from empirical and theoretical perspectives. 

\subsection{Preliminary GANs}
Based on the idea of the zero-sum game, a GAN model consists of a generator $G$ and a discriminator $D$. The generator aims to generate realistic samples to fool the discriminator, while the discriminator tries to distinguish between real and fake samples. When the model reaches the final equilibrium point, the generator will model the target distribution and produce counterfeit samples, which the discriminator will fail to discern. Let $V_D$ and $V_G$ denote the training objectives of the discriminator $D$ and the generator $G$, respectively. The training of the GAN frameworks can be generally illustrated as follows:
\begin{eqnarray}
\max _{D} V_{D} & = & \underset{x \sim {P_{data}}}{\mathbb{E}}\left[D(x)\right]-\underset{z \sim p_z}{\mathbb{E}}\left[D(G(z))\right]
\end{eqnarray}
\begin{eqnarray}
\max _{G} V_{G} & = & \underset{z \sim p_{z}}{\mathbb{E}}\left[D(G(z))\right]
\end{eqnarray}
where the input vector $z$ of $G$ is usually sampled from the \textit{normal distribution}.

\subsection{Motivating Example}

\begin{figure}[htpb]
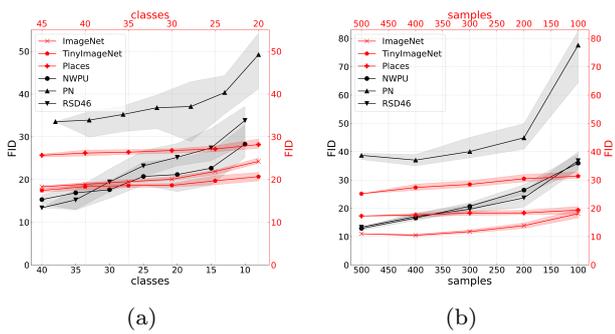

\centering
\subfigure[]{\includegraphics[width=0.48\columnwidth]{newc_f.pdf}}\hspace{2pt}
\subfigure[]{\includegraphics[width=0.48\columnwidth]{news_f.pdf}}
\caption{The FID scores for different experiments on RS datasets : NWPU-RESISC45, PN and RSD46 datasets, and natural dataset: ImageNet Carnivores, TinyImageNet and Places365 datasets. These datasets are trained on StyleGAN2+ADA by varying (a) the number of classes and (b) the number of images per class.}
\label{fig1}
\end{figure}

\begin{figure*}[h]
\centering
\subfigure[]{\includegraphics[width=1\textwidth]{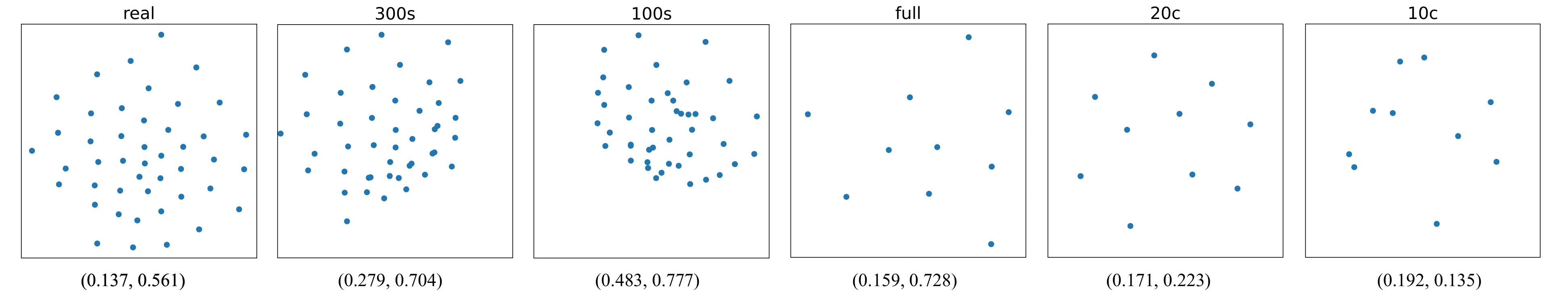}}
\subfigure[]{\includegraphics[width=1\textwidth]{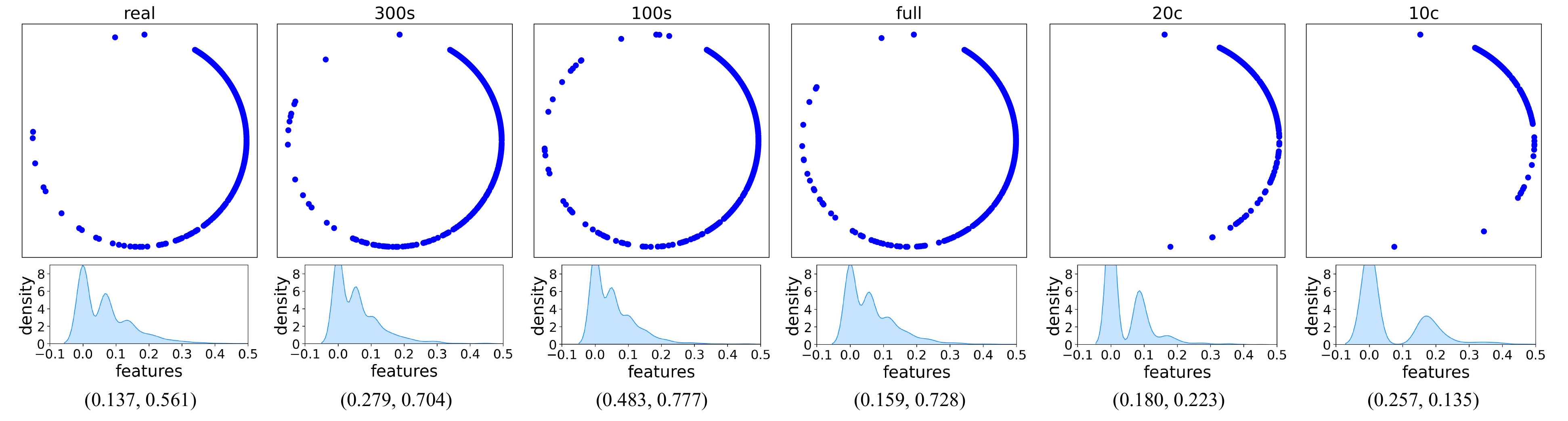}}
\caption{"real" denotes the features extracted from the real dataset. "full" indicates the features extracted from the samples by the GAN model, which is trained under the original dataset. "c" denotes the number of classes used to train the GAN model, and "s" denotes the number of samples. (a) The distributions of samples in the feature space. The samples are generated by GAN models trained under different data setups. We show the cluster centers of samples for simplicity. For fair comparison, we use the ten cluster centers under different class settings. (b) The average feature of samples generated by the same model. We plot the feature distribution on the unit hypersphere $S^1$ and the Gaussian kernel density estimation curves. The two numbers below each chart are the average pairwise $G_2$ potential (the lower, the better) of the distributions and the information entropy (the higher, the better) of the features.}
\label{fig2}
\end{figure*}

\begin{figure*}[htpb]
    \centering
    \subfigure[SimCLR]{\includegraphics[width=1\columnwidth]{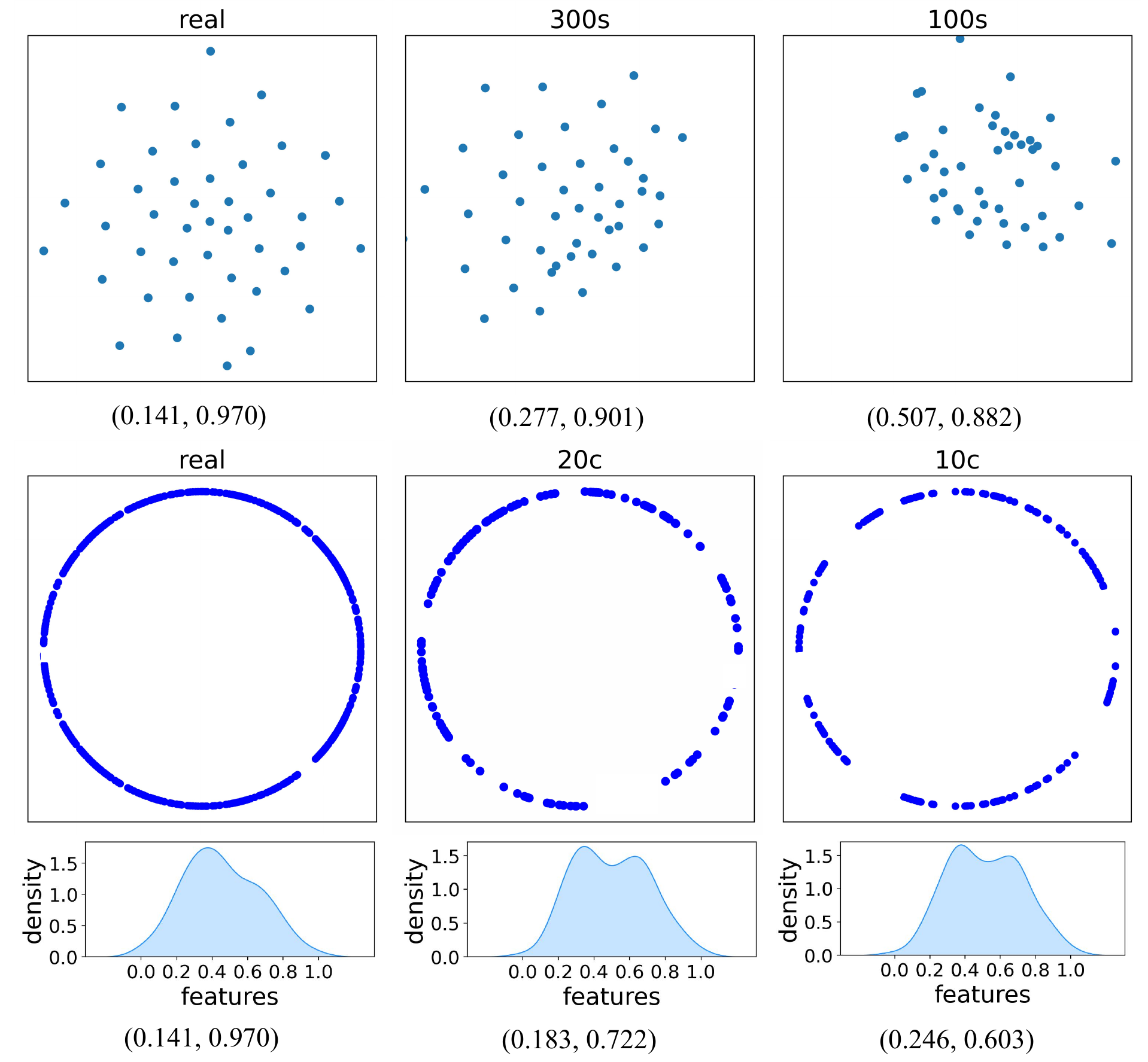}} \hspace{3pt}
    \subfigure[CLIP]{\includegraphics[width=1\columnwidth]{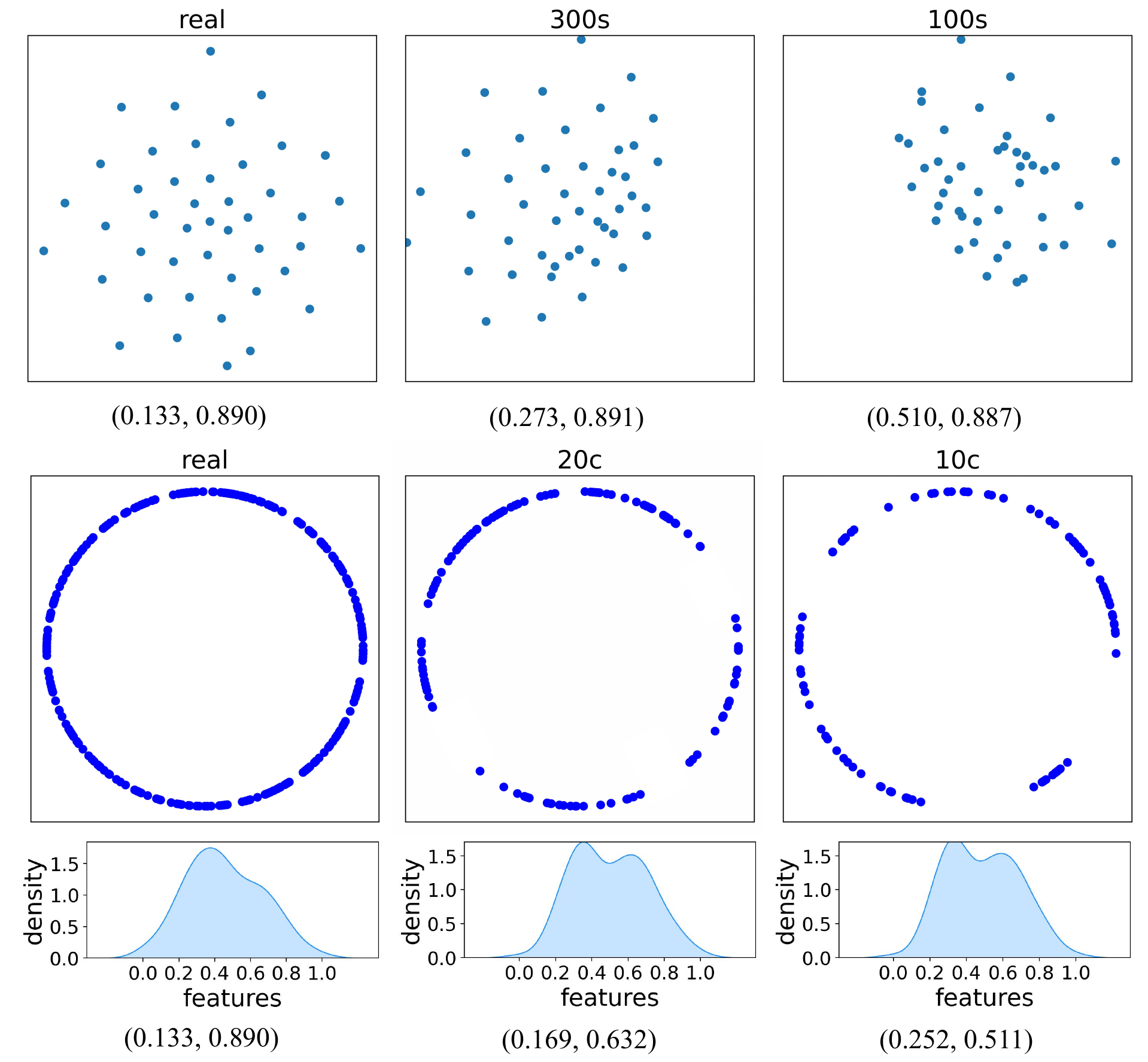}}
    \vspace{10pt}
    \subfigure[ResNet(pretrained)]{\includegraphics[width=1\columnwidth]{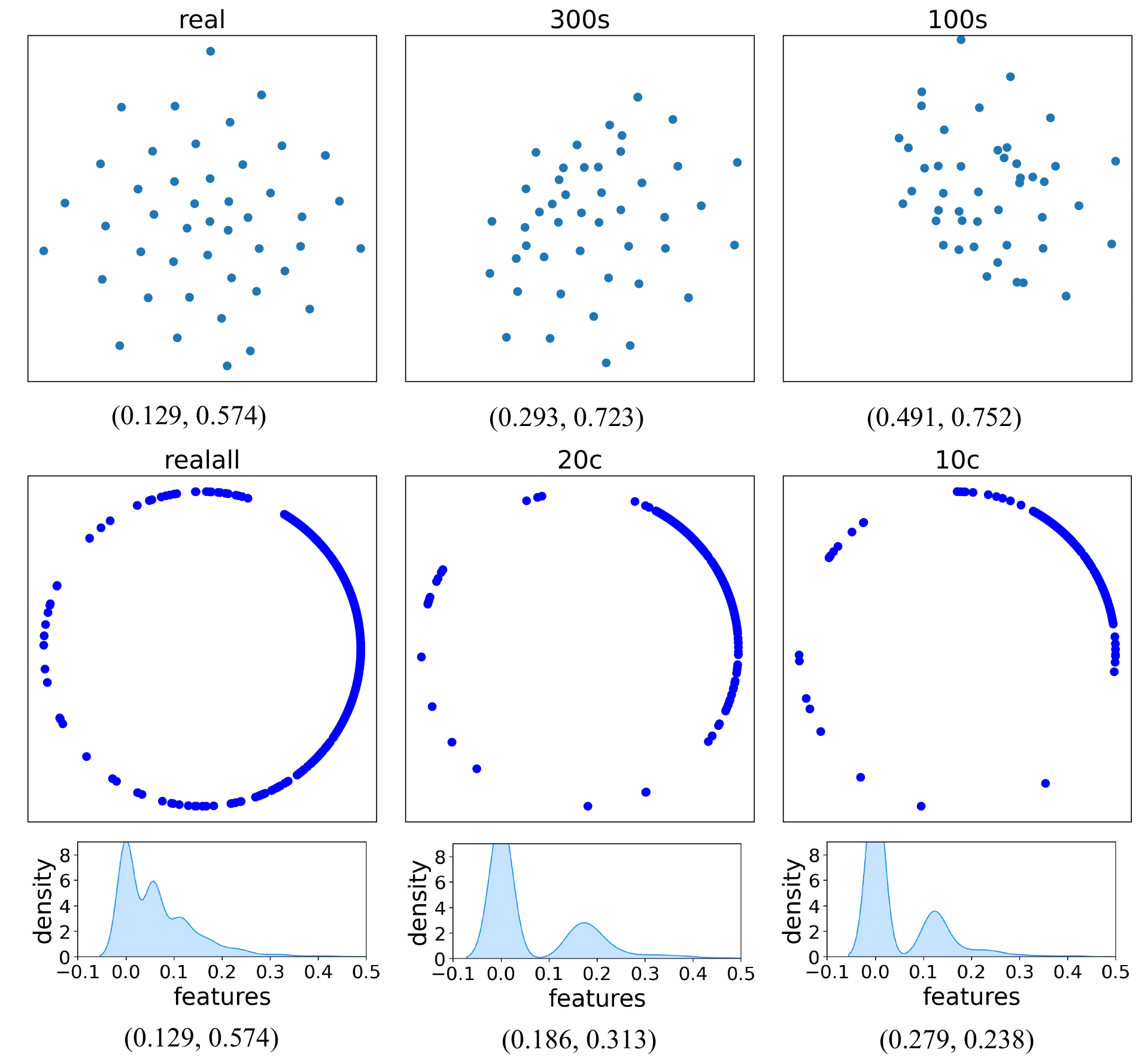}} \hspace{3pt}
    \subfigure[Natural images]{\includegraphics[width=1\columnwidth]{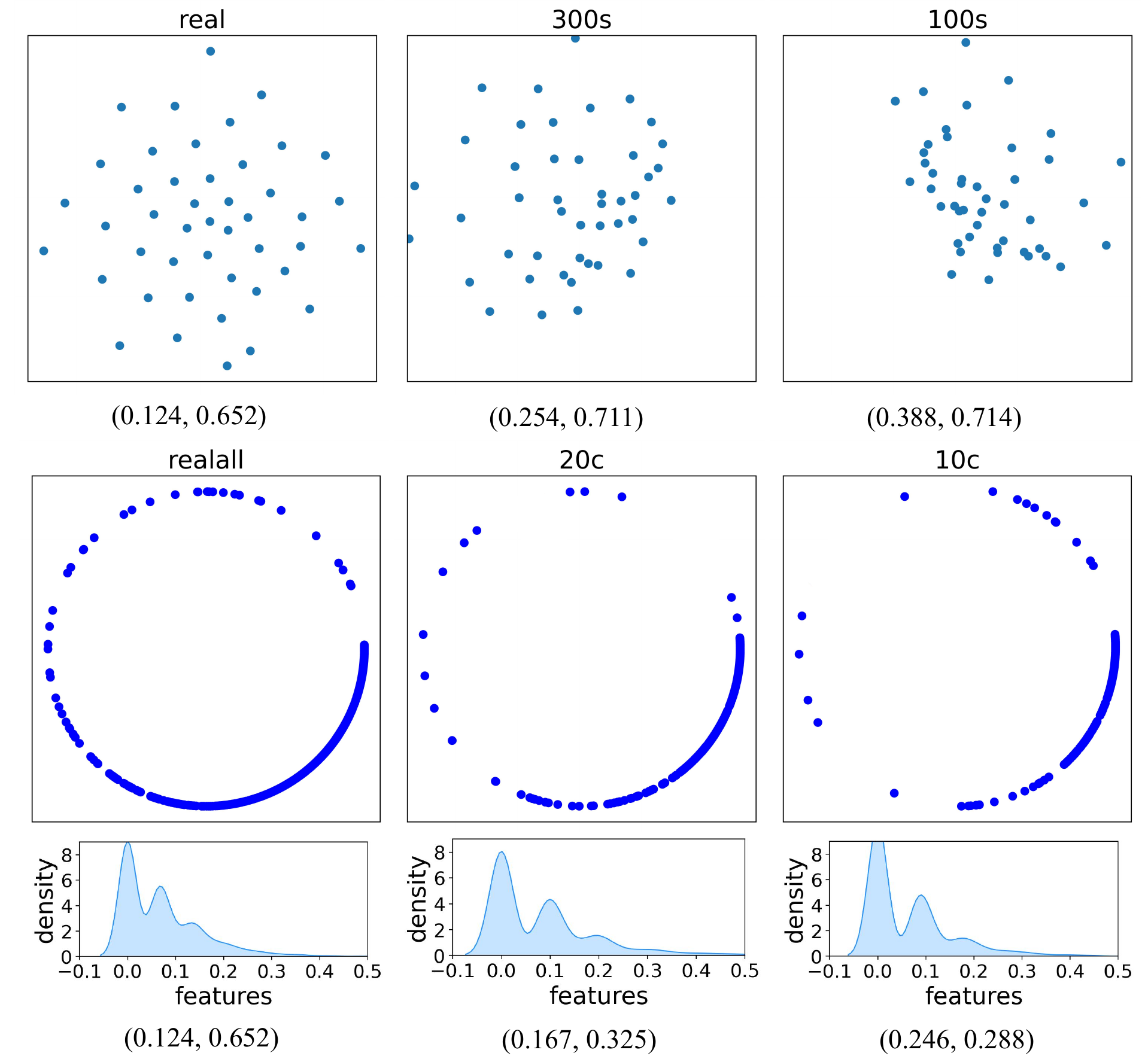}}
\caption{Motivating experiments by different feature extraction networks (a)(b)(c), and on natural images (d).}
\label{fig2-1}
\end{figure*}

As we mentioned above, we apply the well-established GAN models to RS image generation tasks. We test the performance of GANs in the same settings as previous work \citep{transitionCGAN}. We base our experiments on StyleGAN2 \citep{StyleGAN2} with adaptive data augmentation (ADA) \citep{stylegan2-ada}. The model is trained on the NWPU, PN and RSD46 datasets (more details in Section \ref{sec:data}). We perform the experiments by gradually reducing the size of the training set in two ways. In Fig.\ref{fig1}(a), we reduce the number of classes while having 700 training images in each class (800 training images per class in PN dataset, 500 training images per class in RSD46 dataset,). In Fig.\ref{fig1}(b), we reduce the number of images per class while using 45 classes in all cases (38 classes in PN dataset, 40 classes in RSD46 dataset).

To present more comprehensive experimental results, we conduct a minimum of 10 experiments for each experimental setting. For example, in the case of the 20 classes from the NWPU dataset, we randomly select 10 sets of training data, each containing 20 categories, and carry out separate experiments. We compute the average experimental results. In addition, we also present the best and worst results for each experimental setting. In these experiments, we use Frechet inception distance (FID) \citep{FID}, the most commonly-used metric in generative tasks, to measure the quality of generated images. As shown in Fig.\ref{fig1}, the FID gradually increases in both cases, indicating worse generation quality. This observation is in line with previous work \citep{transitionCGAN}. In addition, we found an extraordinary phenomenon: the metric scores of generated RS images vary more significantly than those of natural images under different numbers of training categories or samples per category. In Fig.\ref{fig1}(a), the growth rate of metric scores of RS images are higher than those of natural images (the red curve). Nevertheless, this may be related to the training data size, as there are 700 RS images per category in our experiments, compared to only 100 natural images per category in \citep{transitionCGAN}. But Fig.\ref{fig1}(b) proves that the training data size is not the underlying cause. In Fig.\ref{fig1}(b), we set the number of classes at 45 for NWPU dataset, 38 for PN dataset, and 40 for RSD46 dataset, while the natural images have 50 classes. The growth rate of metric scores of the RS images are still higher than those of the natural images when the samples are reduced.

To demonstrate the generalization of this phenomenon, we conduct additional experiments on two natural datasets: TinyImageNet dataset \citep{tiny-imagenet} and Places365 dataset \citep{zhou2017places}. In our experiments, we randomly choose 50 classes from these two datasets, respectively. And for each class, we randomly choose 500 samples. The experiment results are visualized in Fig.\ref{fig1}. It can be seen that the fluctuations in the FID scores of the RS images are still more drastic than these datasets. It appears that the GAN model is more sensitive to the size of training data for RS image generation tasks than for natural image generation tasks. 

\subsection{Problem Analysis}

For the sensitivity of the GAN model to the size of the RS training dataset, we conjecture that the possible reasons lie in the unique characteristics of RS images compared to natural images. There are two main differences between the two types of images. First, RS images are typically taken from an overhead angle, which captures multiple types of objects such as houses, airplanes, and cars. Second, RS images cover larger areas than natural images. For instance, the NWPU-RESISC45 dataset has a spatial resolution varying from 0.2m to 30m per pixel, while the average scale of a human face is only about 0.2m. These factors result in RS images containing richer feature information than natural images. Therefore, we speculate that the fluctuations of metric scores on RS images may be due to a rapid decrease in the amount of feature information learned by the GAN model as the training data decreases.

In order to confirm our conjecture, we analyze the features of the samples generated under different data settings. To this end, we train ResNet-50 on the NWPU-RESISC45 dataset as the feature extraction model and use the output of the last three blocks of ResNet-50 as the features. The StyleGAN2 is trained under different data setups as shown in Fig.\ref{fig1} and generate the same number of samples in each case. We use the pretrained ResNet-50 to extract features from these samples. It is worth to mention that the ResNet-50 is trained under the corresponding number of classes in the case of reducing classes. 

After obtaining these features, we first conduct dimension reduction by t-SNE\citep{van2008visualizing}
and visualize the distribution of these samples in the feature space in Fig.\ref{fig2}(a). We show the cluster centers of samples for simplicity. For fair comparison, we use forty-five cluster centers under different sample settings (the first row in Fig.\ref{fig2}(a)), and ten cluster centers under different class settings (the second row in Fig.\ref{fig2}(a)). The features from the same setting are computed by the same t-SNE. In other words, the first three figures of Fig.\ref{fig2}(a) are visualized in the same low-dimensionality space, while the others are visualized in the another low-dimensionality space. To present the experimental results more intuitively, they have been visualized separately. It can be observed that the distribution of these samples becomes denser as the sample size per class decreases. In contrast, according to the information theory, the uniform distribution has the maximum entropy, indicating richer information.
Thus, the feature information learned by the GAN model drops with reduced samples. But it seems that the distributions are affected less in the case of reducing classes (the second row in Fig.\ref{fig2}(a)).

Then we analyze the average feature of samples generated from the same model. In order to show the results more intuitively, every element of the average feature is mapped to a 2-dimensional vector on the unit hypersphere $S^{1}$ by sine and cosine functions, and visualized in Fig.\ref{fig2}(b). We also plot the Gaussian kernel density estimation curves of these features under each subfigure.
As we can see visually from Fig.\ref{fig2}(b), the features are affected less when reducing samples, but they get sparser in the case of reducing classes, which means lower information entropy. Thus, the amount of feature information learned by the GAN model also decreases with reduced categories.

To further illustrate our findings, we conduct additional toy experiments based on three different feature-extraction models: self-supervised model SimCLR, CLIP, and ResNet pretrained on ImageNet, which are widely used in downstream tasks.
We also replicate the same experiments as shown in Fig.\ref{fig2} using the ImageNet dataset. The experiment results (Fig.\ref{fig2-1}(a)-(c)) are consistent with Fig.\ref{fig2}. Additionally, the results on natural images (Fig.\ref{fig2-1}(d)) are better than those on RS images. These experimental findings provide further validation for our argument.

From the results of the two experiments above, we empirically prove that the amount of feature information learned by the GAN decreases dramatically as the size of training RS data reduces, leading to poor generation quality. To further investigate this phenomenon, we turn to causal inference theory. 


\begin{figure}[htpb]
\centering
\includegraphics[width=0.8\columnwidth]{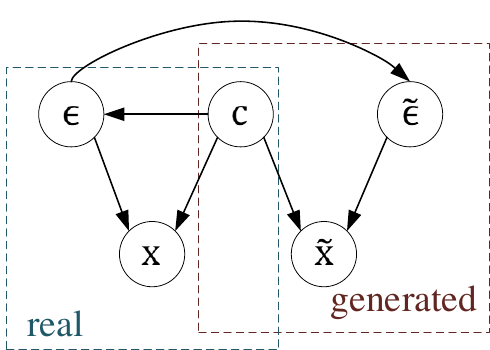} 
\caption{Overview of the image generation process. We partition the latent variable $\mathbf{z}$ into content $\mathbf{c}$ and noise $\mathbf{\epsilon}$. We assume that only noise changes between the real image $\mathbf{x}$ and the generated image $\widetilde{\mathbf{x}}$.}
\label{fig2-2}
\end{figure}

\section{Counterfactual Explanation}
\label{sec:causal}

Let $\mathbf{z}$ be a continuous random variable in a $d$-dimensional representation space $\mathcal{Z} \subseteq {\mathbb{R}^d}$ with associated probability density $p_\mathbf{z}$. Let $\mathbf{x}$ be the continuous random variable in an observation space $\mathcal{X} \subseteq {\mathbb{R}^n}$ which is a low-dimensional sub-manifold of ${\mathbb{R}^n}$. We assume that the real images $\mathbf{x}$ are generated by a smooth function $\mathbf{f}$ which takes a latent code $\mathbf{z}$ as input, thus $\mathbf{x} = \mathbf{f}(\mathbf{z})$. For generated images, we assume that an image $\widetilde{\mathbf{x}}$ is obtained by applying the function $\mathbf{f}$ to a modified representation $\widetilde{\mathbf{z}}$ which is related to the original representation $\mathbf{z}$ of $\mathbf{x}$.

We further assume that the representation $\mathbf{z}$ of images can be partitioned into two disjoint parts:
(i) the content part $\mathbf{c}$, which is shared across $(\mathbf{z}, \widetilde{\mathbf{z}})$, and usually refers to  the high-level structure and semantic information, such as the shape, layout, and object composition;
(ii) the noise part $\mathbf{\epsilon}$, which changes between $(\mathbf{z}, \widetilde{\mathbf{z}})$, and usually refers to the low-level details or random variations of the image, such as texture, color, and lighting.

Intuitively this partition is easy to understand, the generator attempts to learn the real manifold structure and distribution of real images and produce a diverse range of generated images. The generated images have the same high-level content $\mathbf{c}$ as the real images, but different low-level details $\mathbf{\epsilon}$. Since the generator is trained on the real images, the low-level detail $\widetilde{\epsilon}$ is related to the real $\epsilon$. We assume that $\mathbf{c}$ and $\mathbf{\epsilon}$ take values in subspaces $\mathcal{C} \subseteq {\mathbb{R}^{d_c}}$ and $\mathcal{E} \subseteq {\mathbb{R}^{d_{\epsilon}}}$, respectively,i.e., $d = d_c + d_{\epsilon}$ and $\mathcal{Z} = \mathcal{C} \times \mathcal{E}$. Without loss of generality, we let $\mathbf{c}$ corresponds to the first $d_c$ dimensions of $\mathbf{z}$:
\begin{eqnarray}
\mathbf{z} = (\mathbf{c},\mathbf{\epsilon}), \mathbf{c}:= \mathbf{z}_{1:{d_c}},\mathbf{\epsilon}:= \mathbf{z}_{({d_c+1}):d}
\end{eqnarray}

We then establish a structural causal model (SCM) of the above data generating process illustrated in Fig. \ref{fig2-2}. Our approach offers an interpretation of generated data as counterfactuals in the latent SCM. In causal theory, counterfactual is used to analyze and reason about causality and causal relationships by considering what might have happened if circumstances had been different. It involves considering how events or outcomes might have played out if certain variables or factors had been altered. For example, suppose a researcher is studying the effect of a new medication on patient recovery. They might construct counterfactual scenarios by asking questions like, "What would have happened to the patients if they didn't receive the new medication?" or "How would patient recovery differ if they received a placebo instead of the new medication?" In our case, the counterfactual scenario is "What would happen to the image $\mathbf{x}$ if it was generated by $\widetilde{\epsilon }$?". 

We formalize the causal relations as:
\begin{eqnarray}
\mathbf{c}=\mathbf{f}_\mathbf{c}(\mathbf{u}_\mathbf{c}), \mathbf{\epsilon}=\mathbf{f}_\mathbf{\epsilon}(\mathbf{c},\mathbf{u}_\mathbf{\epsilon})
\end{eqnarray}
where $\mathbf{u}_\mathbf{c}$, $\mathbf{u}_\mathbf{\epsilon}$ are independent exogenous variables, and $\mathbf{f}_\mathbf{c}$, $\mathbf{f}_\mathbf{\epsilon}$ are deterministic functions. The latent causal variables $(\mathbf{c}, \mathbf{\epsilon})$ are subsequently decoded into observations $\mathbf{x} = \mathbf{f}(\mathbf{c}, \mathbf{\epsilon})$. Given a factual observation $\mathbf{x}^\mathrm{F}=\mathbf{f}(\mathbf{c}^\mathrm{F}, \mathbf{\epsilon}^\mathrm{F})$ which resulted from $(\mathbf{u}_\mathbf{c}^\mathrm{F},\mathbf{u}_\mathbf{\epsilon}^\mathrm{F})$, the generated data $\widetilde{\mathbf{x}}$ can be interpreted as a result of a soft intervention on $\mathbf{\epsilon}$. This intervention changes $\mathbf{\epsilon}$ the mechanism $\mathbf{f}_\mathbf{\epsilon}$ to $do(\mathbf{\epsilon}:=\widetilde{\mathbf{f}}_\mathbf{\epsilon}(\mathbf{c},\mathbf{u}_\mathbf{\epsilon},\mathbf{u}_A))$, where $\mathbf{u}_A$ accounts for the randomness of the generation process. The generated data can be referred as the counterfactual observations 
$\mathbf{x}^\mathrm{CF}=\mathbf{f}(\mathbf{c}^\mathrm{F},\mathbf{\epsilon}^\mathrm{CF})$.

Our objective is to demonstrate the correlation between the feature information learned by the GAN model and the quality of generated images. In image generation tasks, if the generator could learn superior content information $\mathbf{c}$ that are closed to those of real images, the quality of the generated images would enhance. We start with the following definition.

\begin{definition}
    The true content partition $\mathbf{c}=\mathbf{f}^{-1}(\mathbf{x})_{1:d_c}$ is block-identified by a function $\mathbf{g}:\mathcal{X} \rightarrow \mathcal{Z}$ if the inferred content partition $\widehat{\mathbf{c}}=\mathbf{g}(\mathbf{x})_{1:d_c}$ contains all and only information about $\mathbf{c}$, i.e., if there exists an invertible function $\mathbf{h}:\mathbb{R}^{d_c} \rightarrow \mathbb{R}^{d_c}$ s.t. $\widehat{\mathbf{c}}=\mathbf{h}(\mathbf{c})$.
\end{definition}

We assume that if the representations of the generated images could be block identified by any smooth function $\mathbf{g}$, then the generator learn the true content variable.

\begin{theorem}
    Consider the data generating process described in Fig.\ref{fig2-2}, and assume further that:

    (i) $\mathbf{f}:\mathcal{Z} \rightarrow \mathcal{X}$ is smooth and differentiable. 
    
    (ii) $p_\mathbf{z}$ is a smooth, continuous density on $Z$ with $p_\mathbf{z}(\mathbf{z}) > 0$ almost everywhere;
    
    (iii) for any $l \in {1, ..., d_{\epsilon}}$, $\exists A \subseteq {1, ..., d_{\epsilon}}$ s.t. $l\in A$; $p_A(A) > 0$; $p_{\widetilde{\mathbf{\epsilon}}_A|\mathbf{\epsilon}_A}$ is smooth w.r.t. both $\mathbf{\epsilon}_A$ and $\widetilde{\mathbf{\epsilon}}_A$; and for any $\mathbf{\epsilon}_A$, $p_{\widetilde{\mathbf{\epsilon}}_A|\mathbf{\epsilon}_A}(\cdot|\mathbf{\epsilon}_A)>0$ in some open, non-empty subset containing $\mathbf{\epsilon}_A$.
    
    Let $\mathbf{g}:\mathcal{X}\rightarrow (0, 1)_{d_c}$ be any smooth function which minimises the following functional:
    \begin{footnotesize}
    \begin{equation}
    \begin{aligned}    
       \mathcal{L}_{\text {AlignMax}}(\mathbf{g}):= \mathbb{E}_{(\mathbf{x}, \tilde{\mathbf{x}}) \sim p_{\mathbf{x}, \tilde{\mathbf{x}}}}\left[\|\mathbf{g}(\mathbf{x})-\mathbf{g}(\tilde{\mathbf{x}})\|_{2} \|_{2}^{2}\right]-H(\mathbf{g}(\mathbf{x})) \nonumber
    \end{aligned}
    \end{equation}
    \end{footnotesize}
    where $H(\cdot)$ denotes the differential entropy of the random variable $\mathbf{g}(\mathbf{x})$ taking values in $(0, 1)_{d_c}$.Then $\mathbf{g}$ block-identifies the true content variables.
\end{theorem}

The proof of \textit{Theorem} 1 can be found in the Appendix and comprises three main steps. In the first step, we demonstrate that the representation $\hat{\mathbf{c}}=\mathbf{g}(\mathbf{x})$ extracted by any smooth function $\mathbf{g}$ that minimises $\mathcal{L}_{\text{AlignMax}}$ is linked to the true latent $\mathbf{z}$ through a smooth mapping $\mathbf{h}$. This mapping ensures that $\hat{\mathbf{c}}$ must exhibit invariance across $(\mathbf{x}, \tilde{\mathbf{x}})$ almost surely with respect to the true generative process $p$, and that $\hat{\mathbf{c}}$ must follow a uniform distribution on $(0,1)^{d_{c}}$. In the second step, we then use assumptions (ii) and (iii) to demonstrate, through contradiction, that  $\hat{\mathbf{c}}=\mathbf{h}(\mathbf{z}) = \mathbf{h}(\mathbf{c}, \mathbf{\epsilon})$ can only depend on the true content $\mathbf{c}$ and not on the noise variable $\epsilon$. Otherwise, the invariance established in the first step would be violated with probability greater than zero. Finally, in the third step, we prove that $\mathbf{h}$ must be a bijection, i.e., invertible.

\begin{figure*}[htpb]
\centering
\includegraphics[width=0.9\textwidth]{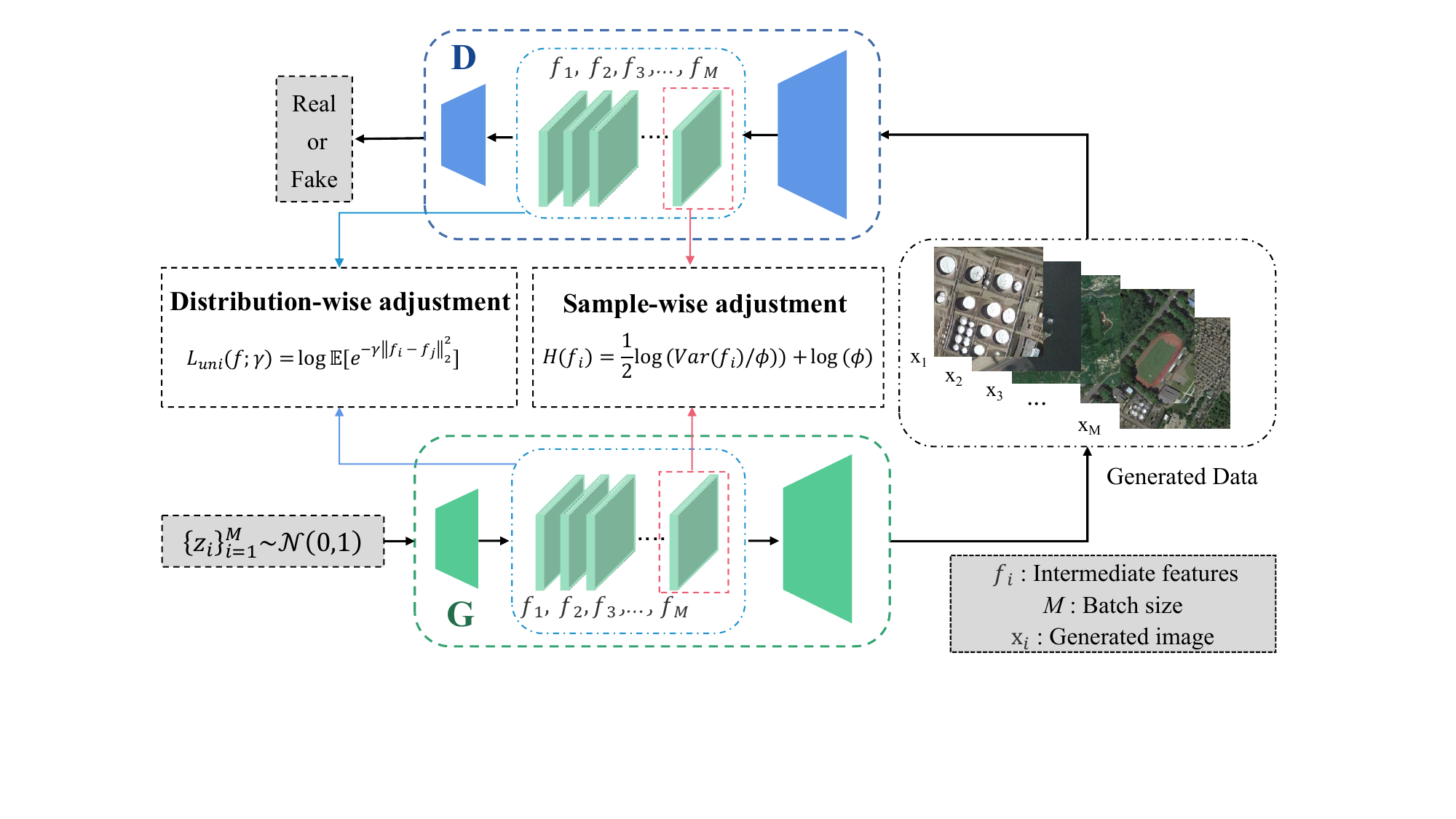} 
\caption{The overall architecture of our method. $G$ and $D$ are the generator and the discriminator. $f$ denotes the intermediate features. $M$ is the batch size. The real dataset is omitted for clarity as our method is only imposed on generated data.}
\label{fig3}
\end{figure*}

The first term in $\mathcal{L}_{\text{AlignMax}}$ represents the original objective of the GAN generator, while the second term is the entropy of the learned representations. \textit{Theorem} 1 states that maximizing the entropy of the learned representation can recover the true content variable of real images. This is intuitively valid because any function solely dependent on $\mathbf{c}$ will remain invariant across $(\mathbf{x}, \tilde{\mathbf{x}})$, so it is beneficial to preserve all content information to maximize entropy. This theorem initially proves the entropy of the learned representations is positively correlated with the quality of the generated images. It subsequently provides insight into enhancing the quality of generated images by enriching feature information. Recall that the distribution of samples in the feature space gets denser when reducing samples per class, and the features of individual samples become sparser with reduced classes. Consequently, we design two novel adjustment schemes that increase the entropy of the representations learned by the GAN model at the distribution and sample levels, which will be described in detail below.


\section{Methods}
\label{sec:methods}
This paper proposes to increase the the amount of feature information contained in the GAN model from two aspects, distribution-wise adjustment and sample-wise adjustment. The overall architecture is shown in Fig.\ref{fig3}. Our methods require neither modification of the network structure nor excessive computational effort and can be used for arbitrary generative models. Next, we will introduce them in detail.

\subsection{Distribution-wise Adjustment}
\label{sec:3.1}

In this section, we aim to design distribution-wise adjustment to enrich the feature information learned by the GAN model. According to the definition of information entropy, the greater the entropy is, the more information it contains. It is well known that the uniform distribution has the greatest information entropy. Intuitively, vectors that are roughly uniformly distributed on the unit hypersphere could preserve more information than other vectors. Consequently, we aim to design a novel technique named Uniformity Regularization (UR) over feature space to constrain the feature distribution to a uniform distribution. Designing such a regularization term is in fact nontrivial, it should meet two requirements. To start with, the regularization term should be asymptotically correct, i.e., the distribution optimizing this metric should converge to uniform distribution. Then, the regularization term is supposed to be empirically reasonable with finite number of points. To this end, we consider the Gaussian potential kernel.
\begin{eqnarray}
G(x, y) \triangleq e^{-\gamma\|x-y\|_{2}^{2}} & = & e^{2 \gamma \cdot x^{\top} y\
-2 \gamma}, \quad \gamma>0
\label{equation1}
\end{eqnarray}

In contrast to other kernels that achieve uniformity of optimal points, the Gaussian kernel is closely related to the universal optimal point configuration and can also be used to represent a general class of other kernels \citep{gaussianK}. Therefore, the uniformity regularization term is based on Gaussian potential kernel and can be viewed as the logarithm of the average pairwise Gaussian potential, which is shown as follows:
\begin{eqnarray}
\label{equation2}
\mathcal{L}_{\text {uni }}(f ; \gamma) & =  \log \underset{\substack{\text { i.i.d. } \\
x, y \sim p_{\text {G }}}}{\mathbb{E}}\left[e^{-\gamma \|f(x)-f(y)\|_{2}^{2}}\right]
\end{eqnarray}
where $P_G$ denotes the distribution of samples generated by the generator $G$, $f$ denotes the intermediate layer of generator or discriminator, and $\gamma>0$ is a hyperparameter.
\begin{theorem}
The distribution that minimize the pairwise Gaussian potential is the uniform distribution ${\sigma}_d$ on the unit hypersphere $S_d$. 
\end{theorem}

\begin{theorem}
As number of points reaches infinity, distributions of points minimizing the average pairwise potential converge weak* to the uniform distribution ${\sigma}_d$. 
\end{theorem}

The proofs of \textit{Theorem} 2 and \textit{Theorem} 3 can be seen in the Appendix. In accordance with \textit{Theorem} 2, it is evident that the uniform distribution stands as the sole distribution minimizing the expected pairwise potential. In line with \textit{Theorem} 3, we can conclude that, with a finite number of points, the pairwise Gaussian potential is empirically justifiable. Based on these two theorems, the uniformity regularization term satisfies both of the aforementioned requirements. Besides, we evaluate the average pairwise potential of various finite point collections on $S^1$ in Fig.\ref{fig2}(a). The values are consistent with our intuitive understanding of uniformity. 

It has been shown that different network layers are responsible for different levels of detail in the generated image. Empirically, the latter blocks of the generator have more effect on the style (e.g. texture and color) of the image whereas the earlier blocks impact the coarse structure or content of the image \citep{styleclip}. Thus, we choose features from a shallow network layer for adjustment in our experiments.

\subsection{Sample-wise Adjustment}
\label{sec:3.2}

While previous section focuses on distribution-wise adjustment, this section designs a novel approach at the sample level to increase the amount of feature information learned by the GAN model. Inspired by the principle of maximum entropy in information theory, we propose sample-wise adjustment to increase the entropy of the features from a single generated sample. To this end, we design an Entropy Regularization (ER) term to calculate entropy of the intermediate features $f(x)$. 

Since the feature distribution is a high dimensional function, it is difficult to compute the precise value of its entropy directly. Fortunately, based on the \textit{Theorem} 4 \citep{information}, we could estimate the upper bound of the entropy.
\begin{theorem}
For any continuous distribution $P(x)$ of mean ${\mu}$ and variance ${\sigma}^2$, its differential entropy is maximized when $P(x)$ is a Gaussian distribution $N({\mu},{\sigma}^2)$.
\end{theorem}

According to \textit{Theorem} 4, the entropy of a distribution is upper bounded by a Gaussian distribution with the same mean and variance. The entropy of a Gaussian distribution is shown in Eq.\ref{equation3}:
\begin{eqnarray}
\label{equation3}
H^{*}(x) & = & \frac{1}{2} \log (2 \pi)+\frac{1}{2}+H(x) \\
H(x): & = & \log (\sigma) \nonumber
\end{eqnarray}
where $x$ is sampled from the Gaussian distribution $N({\mu},{\sigma}^2)$.

Since the entropy of Gaussian distribution only depends on the variance, we could use $H(x)$ instead of $H^{*}(x)$. Based on these discussions, the upper bound entropy of the intermediate features $f(x)$ is shown in Eq.\ref{equation4}:
\begin{eqnarray}
\label{equation4}
H(f(x)) & = & \frac{1}{2} \log \left(\operatorname{Var}\left(f(x)\right)\right)
\end{eqnarray}
where $x$ is a generated image.

To avoid numerical overflow, we re-scale features $f(x)$ by a constant ${\phi}$, and then compensate the entropy of the features by multiplying this constant. In our experiments, we set ${\phi}$ to the L2 norm of the features. We evaluate the entropy of the features in Fig.\ref{fig2}(b). The results prove the efficacy of our method. In addition, same as previous section, we choose features from a shallow network layer for adjustment. 
\begin{eqnarray}
\label{equation5}
H(f(x)) & = & \frac{1}{2} \log \left(\operatorname{Var}\left(f(x)/\phi \right)\right)+ \log \left(\phi \right)
\end{eqnarray}

\subsection{Overall Objective}
\label{sec:3.3}

The final loss function of the generator is shown in Eq.\ref{equation6}:
\begin{eqnarray}
\label{equation6}
\begin{aligned}
&& {L}_{\mathrm{G}} & = {L}_{\text{ori}}+\lambda_{\mathrm{G}} {L}_{\text{uni}}^{\mathrm{G}}
+\lambda_{\mathrm{D}} {L}_{\text{uni}}^{\mathrm{D}} & \\
&& \ & -\delta_{\mathrm{G}} \sum {H}^{\mathrm{G}}
-\delta_{\mathrm{D}} \sum {H}^{\mathrm{D}}
\end{aligned}
\end{eqnarray}
where $L_{\text{ori}}$ represents the original loss function of generator. $L_{\text{uni}}^G$ and $L_{\text{uni}}^D$ denote the uniformity regularization on features from the generator and the discriminator respectively. $H^G$ and $H^D$ denote the entropy regularization on features from the generator and the discriminator respectively. ${\lambda}_G$, ${\lambda}_D$, ${\delta}_G$, and ${\delta}_D$ are hyperparameters.

\begin{figure*}[h]
\centering
\subfigure[]{\includegraphics[width=1\textwidth]{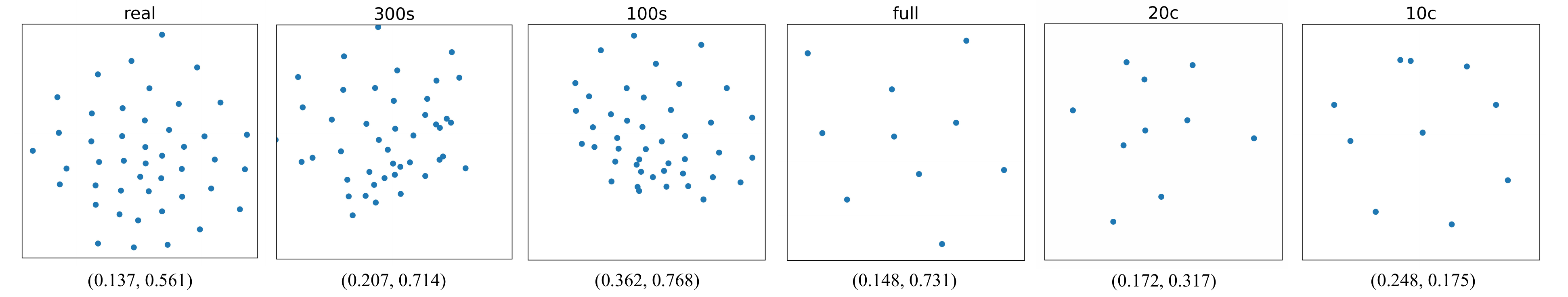}}
\subfigure[]{\includegraphics[width=1\textwidth]{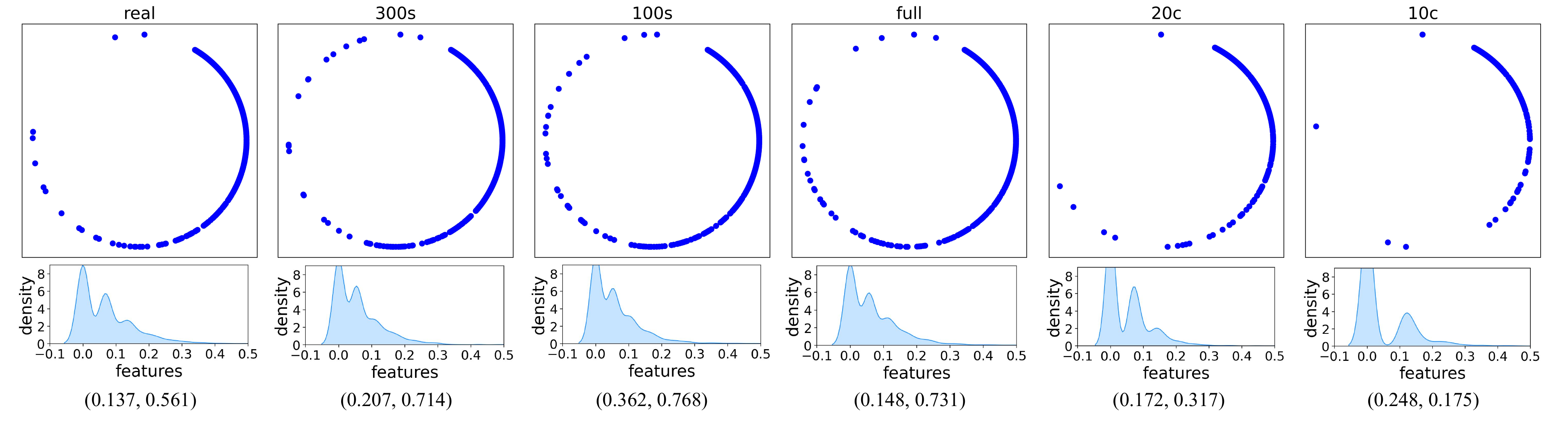}}
\caption{(a) The distributions of samples generated by our methods in the feature space. (b) The average feature of samples generated by the same model. The two numbers below each chart are the average pairwise $G_2$ potential (the lower, the better) and the information entropy (the higher, the better).}
\label{fig4}
\end{figure*}

\section{Experiment}
\label{sec:ex}
In this section, we first provide the details of our experimental setup. Then, we present the quantitative and qualitative results of the proposed method, as well as the comparison with existing methods. Finally, we provide more ablation and analysis of different components of our method. In our experiments, "real" denotes real data. "full" indicates the original dataset. "c" denotes the number of classes used to train the GAN model, and "s" denotes the number of samples.
\subsection{Experiment Setup}
\label{sec:data}
\textbf{Datasets}: We use four remote sensing datasets to evaluate our method: NWPU-RESISC45 Dataset, PatternNet Dataset, RSD46-WHU Dataset and AID Dataset. To evaluate our methods under different data set up, we decrease the number of classes and images per class in these datasets using random sampling.

The NWPU-RESISC45 Dataset \citep{NWPU} has 31,500 images covering more than 100 countries and regions around the world. It has 45 categories with 700 images in each category. Each image is 256×256 pixels in size. The spatial resolution of this dataset is up to 0.2m and the lowest is 30m. In addition, the images are varied in lighting, shooting angle, imaging conditions, and so on. 

The PatternNet Dataset \citep{patternnet} is a large-scale high-resolution remote sensing dataset. It has 38 categories with 800 images in each category. Each image is 256×256 pixels in size. The spatial resolution of this dataset varies from 0.06m per pixel to 4.7m per pixel. The images in PatternNet are collected from Google Earth imagery or via the Google Map API for some US cities.  

The AID Dataset \citep{AID} has 10,000 images and each image is 600×600 pixels in size. The images are multi-source since they are collected from Google Earth imagery which contains different remote sensing sensors. It has 30 categories and there are about 220–420 images in each category. In our experiments, these images are resized to 256×256.

The RSD46-WHU Dataset \citep{long2017accurate} is a large-scale open dataset for scene classification in RS images, collected from Google Earth and Tianditu. Most classes in the dataset have a ground resolution of 0.5m, with others at around 2m. The dataset includes 500-3000 images in each class, totaling 117,000 images across 46 classes. In our experiments, we selected 40 classes, each containing 500 images.

\textbf{Implementation details}: We base our method on two different models \textbf{StyleGAN2} \citep{StyleGAN2} and \textbf{BigGAN} \citep{biggan} with different augmentation methods \textbf{ADA} \citep{stylegan2-ada} and \textbf{APA} \citep{APA}, and regularization method \textbf{LeCam} \citep{lecam}. We use the official PyTorch implementation of StyleGAN2+ADA. For StyleGAN2+APA, BigGAN and LeCam, we use the implementations provided by \citep{contragan}.

\textbf{Evaluation Metrics}: We evaluate our method using Frechet inception distance (FID) \citep{FID}, as the most commonly-used metric for measuring the quality and diversity of images generated by GAN models. We also include Kernel Inception Distance (KID) \citep{KID} as a metric that is unbiased by empirical bias \citep{xu2018empirical}.

\subsection{Results and Comparisons}

\label{sec:res}
\begin{table*}[htb]%
\centering
\caption{Comparison of Quality Scores (FID, KIDx${10}^{3}$) on the NWPU-RESISC45, PatternNet and AID Datasets (the \textcolor{red}{red} numbers present our improvement)}
\renewcommand\arraystretch{1.5}
\begin{tabular}{lllllll}
\hline
\multicolumn{1}{c}{\multirow{2}{*}{\textbf{Methods}}} & \multicolumn{2}{c}{\textbf{NWPU}}                 & \multicolumn{2}{c}{\textbf{PN}}                    & \multicolumn{2}{c}{\textbf{AID}}                           \\ \cline{2-7} 
\multicolumn{1}{c}{}                         & \multicolumn{1}{c}{\textbf{FID} ${\downarrow}$} & \multicolumn{1}{c}{\textbf{KID${\downarrow}$}} & \multicolumn{1}{c}{\textbf{FID}${\downarrow}$} & \multicolumn{1}{c}{\textbf{KID}${\downarrow}$} & \multicolumn{1}{c}{\textbf{FID}${\downarrow}$} & \multicolumn{1}{c}{\textbf{KID}${\downarrow}$} \\ \hline
BigGAN+ADA                                   & 30.91\textcolor{red}{-2.09}              & 10.13\textcolor{red}{-1.41}              & 43.94\textcolor{red}{-1.59}              & 18.71\textcolor{red}{-1.03}              & 42.26\textcolor{red}{-2.04}              & 18.08\textcolor{red}{-1.52}              \\
BigGAN+ADA+LeCam                             & 32.65\textcolor{red}{-2.24}              & 11.69\textcolor{red}{-1.65}              & 41.28\textcolor{red}{-3.42}              & 17.89\textcolor{red}{-1.97}              & 38.23\textcolor{red}{-2.23}              & 15.44\textcolor{red}{-1.67}              \\
StyleGAN2+APA                                & 21.67\textcolor{red}{-2.91}              & 6.54\textcolor{red}{-1.90}               & 33.75\textcolor{red}{-1.84}              & 12.34\textcolor{red}{-2.33}              & 32.51\textcolor{red}{-3.22}              & 11.73\textcolor{red}{-2.05}              \\
StyleGAN2+ADA                                & 11.97\textcolor{red}{-2.23}              & 3.27\textcolor{red}{-1.82}               & 33.53\textcolor{red}{-2.06}              & 12.67\textcolor{red}{-1.92}              & 27.83\textcolor{red}{-3.42}              & 7.52\textcolor{red}{-1.73}               \\
StyleGAN2+ADA+LeCam                          & 14.38\textcolor{red}{-2.68}              & 4.97\textcolor{red}{-1.39}               & 31.32\textcolor{red}{-2.11}              & 11.32\textcolor{red}{-0.89}              & 28.63\textcolor{red}{-2.37}              & 7.79\textcolor{red}{-1.88}               \\ \hline
\end{tabular}
\label{table1}
\end{table*}

\begin{figure*}[htpb]
\centering
\subfigure[]{\includegraphics[width=0.185\textwidth]{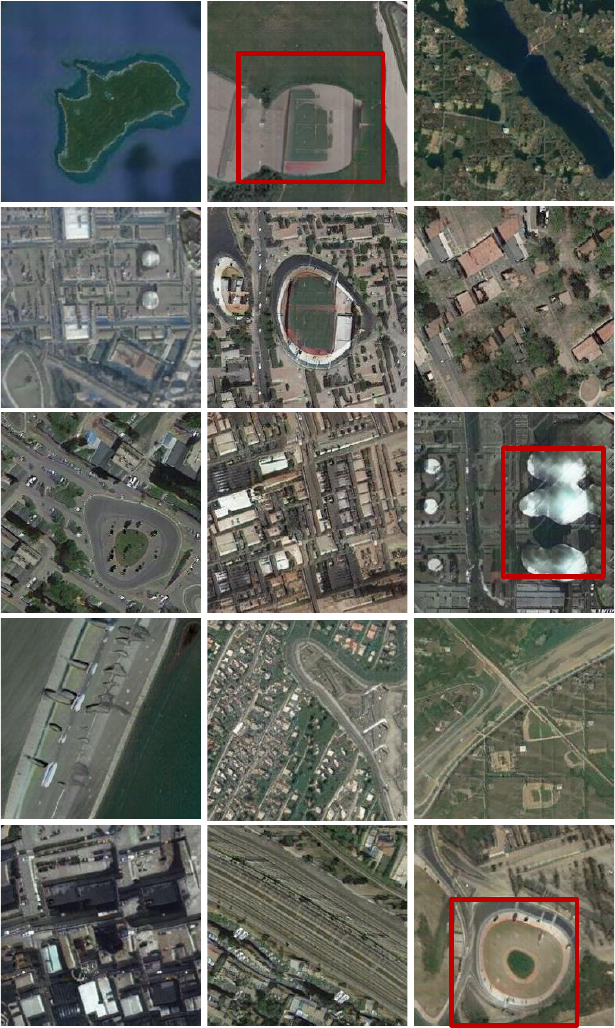}} \hspace{1pt}
\subfigure[]{\includegraphics[width=0.185\textwidth]{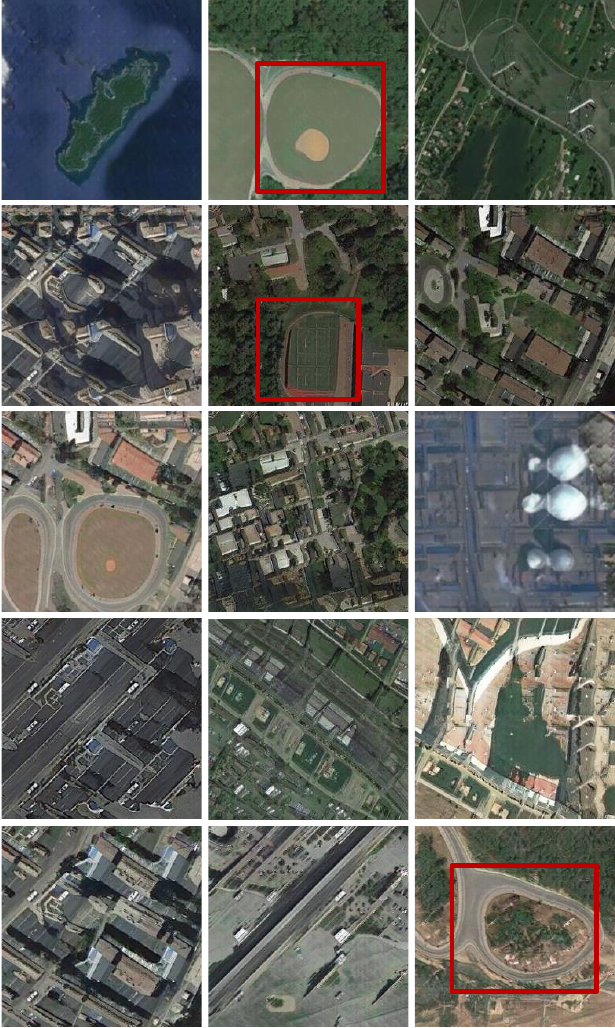}} \hspace{1pt}
\subfigure[]{\includegraphics[width=0.185\textwidth]{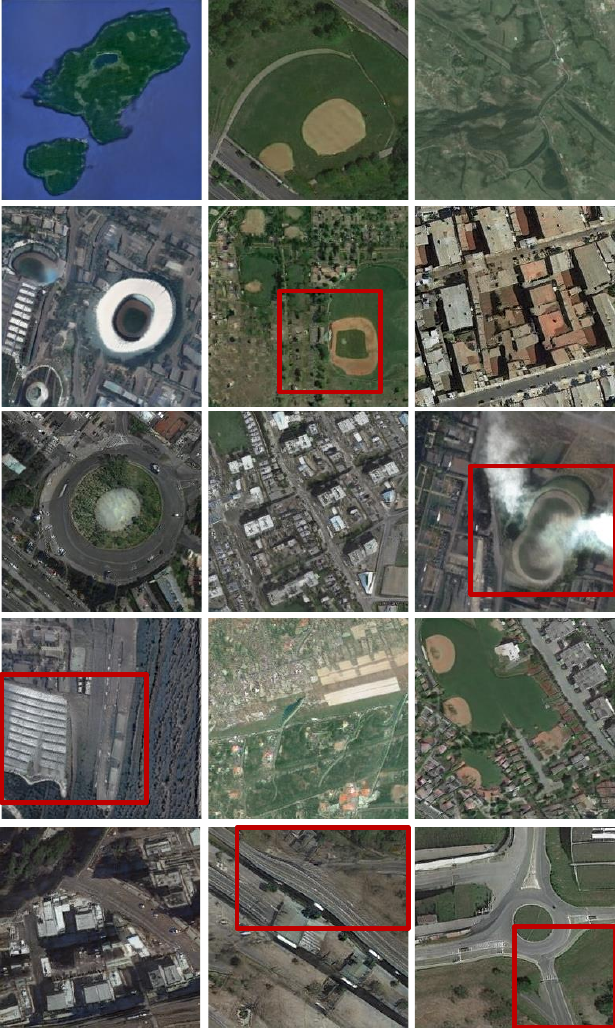}} \hspace{1pt}
\subfigure[]{\includegraphics[width=0.185\textwidth]{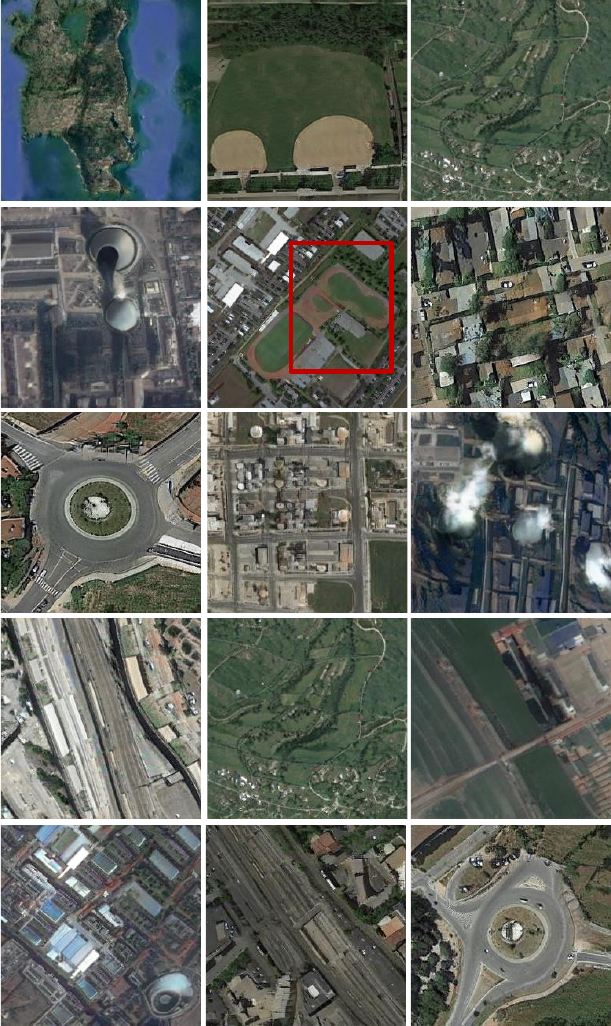}} \hspace{1pt}
\subfigure[]{\includegraphics[width=0.185\textwidth]{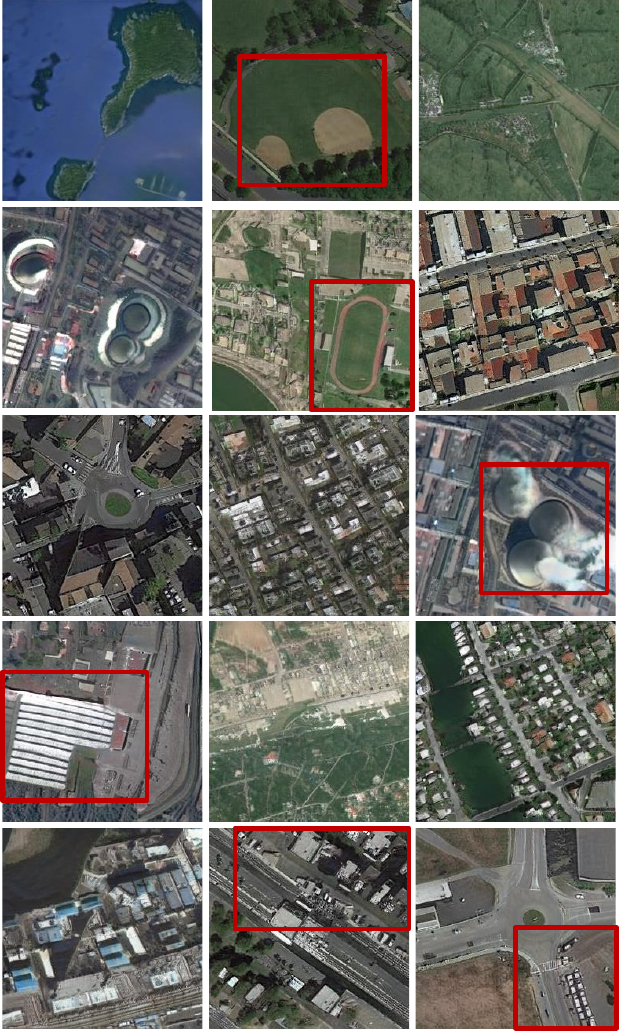}}
\caption{The generated images of different methods on NWPU dataset. (a) BigGAN+ADA. (b) BigGAN+ADA+Lecam. (c) StyleGAN2+ADA. (d) StyleGAN2+ADA+Lecam. (e) Our method.}
\label{fig_new_NWPU}
\end{figure*}

\begin{figure}[htpb]
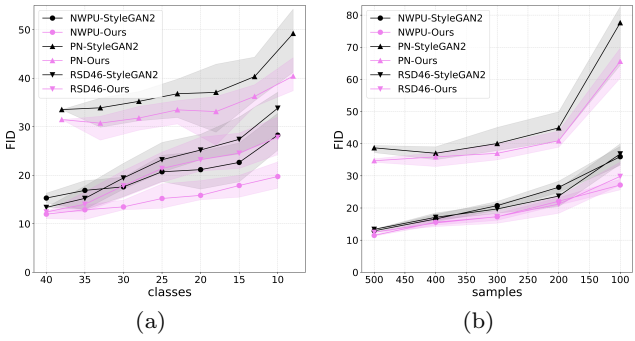

\centering
\subfigure[]{\includegraphics[width=0.48\columnwidth]{newc_fm.pdf}}
\hspace{2pt}
\subfigure[]{\includegraphics[width=0.48\columnwidth]{news_fm.pdf}}
\caption{The FID curves of our method on NWPU, PN and RSD46 datasets, by varying (a) the number of classes and (b) the number of images per class.}
\label{fig1-1}
\end{figure}

We first visualize the generated images on NWPU in Fig.\ref{fig_new_NWPU}. The images located at corresponding positions across different models in these figures belong to the same category. From these figures, we can observe that the foreground content of the images generated by our method is more detailed and exhibits a higher degree of realism, while the shapes of objects are more consistent and regular. These aspects are highlighted using red boxes. More visual results on NWPU, PN and AID datasets are available in the Appendix. We also visualize the distributions and features of our method in Fig.\ref{fig4}. Compared with Fig.\ref{fig2}, the distributions of generated images in the feature space under fewer training samples become more uniform (Fig.\ref{fig4}(a)), and the features of the individual images under fewer training classes are denser (Fig.\ref{fig4}(b)). As shown in Fig.\ref{fig1-1}, our method performs well under various data settings and mitigates the upward trend of the metric curves. The experiment results prove the efficacy of our methods. 

\begin{figure}[htpb]
\centering
\subfigure[]{\includegraphics[width=0.7\columnwidth]{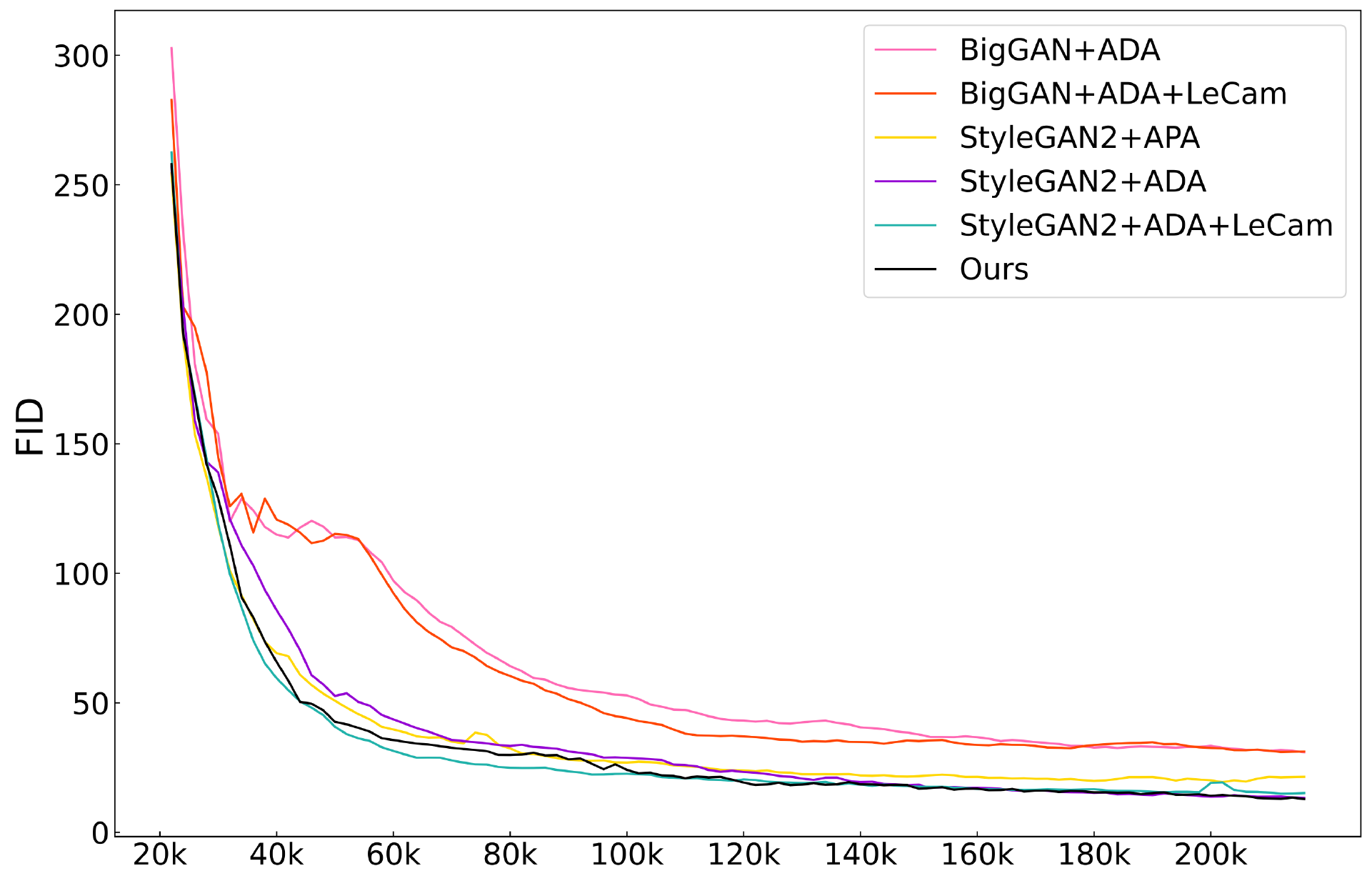}}\hspace{5pt}
\subfigure[]{\includegraphics[width=0.7\columnwidth]{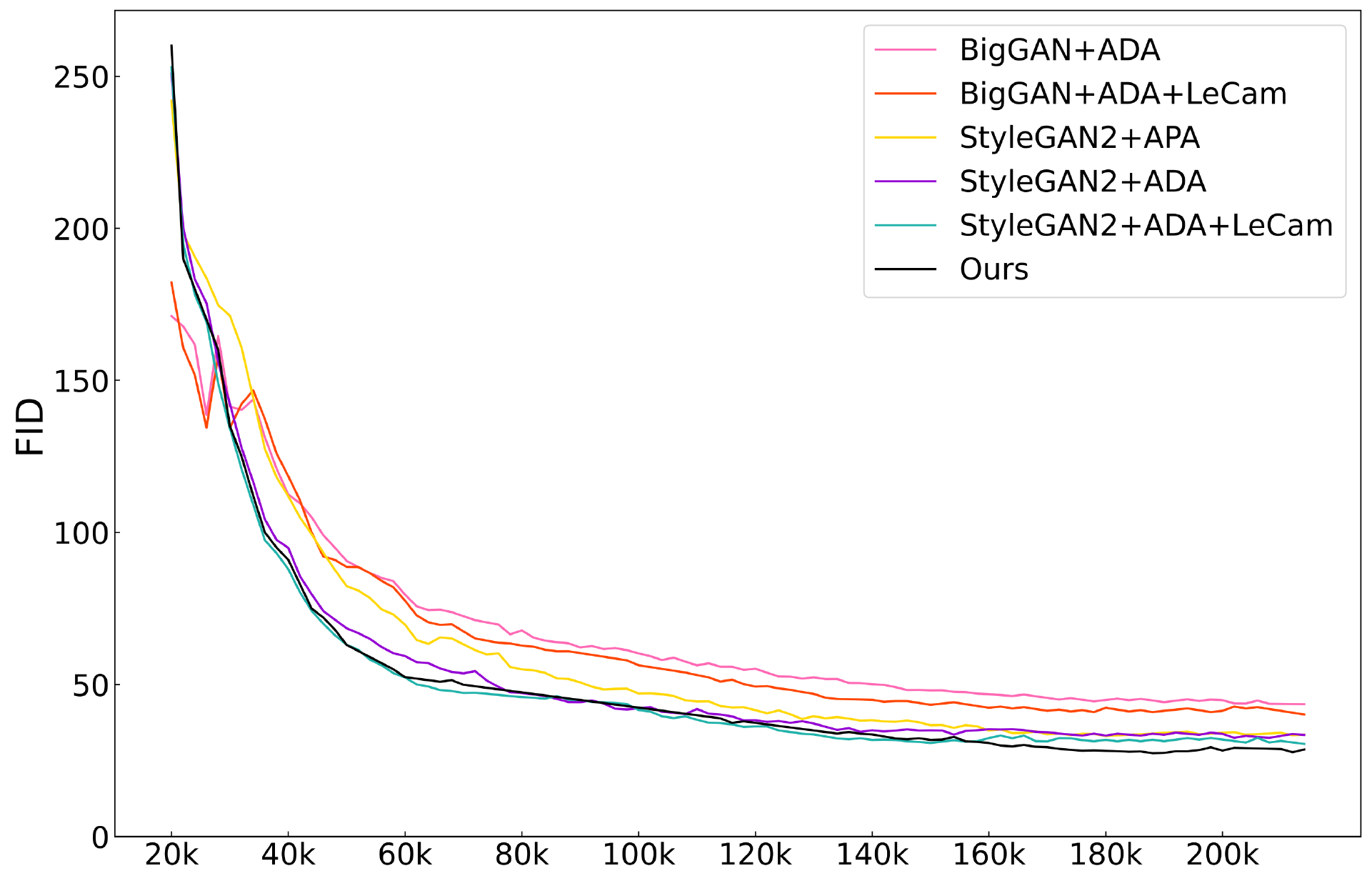}}
\caption{The FID curves on the (a) NWPU-RESISC45 dataset and (b) PatternNet dataset.}
\label{fig5}
\end{figure}

Then, we provide a quantitative comparison with the well-established baselines. Results are reported in Table \ref{table1}. The red numbers indicate the improvement of the GAN models after using our method. In general, the FID and KID scores for our proposed method indicate a significant and consistent advantage over all the compared methods. In detail, our comparison experiments can be divided into three types. First, under different model structures, BigGAN and StyleGAN2, our method is robust, which proves that our approach can be applied to any model architecture. Second, under different augmentation methods, ADA and APA. The results of ADA are better than those of APA on RS datasets. Nevertheless, our method still performs well. This shows that the proposed regularization approaches can be applied to other GAN models along with existing augmentation approaches. Third, combined with the regularization method LeCam, our method is still effective. The proposed method can be viewed as an effective complement to existing regularization methods. The FID curves of StyleGAN2 and BigGAN on NWPU and PN datasets with different optimized approaches during training are visualized in Fig.\ref{fig5}. Our method achieves stable training dynamics and better FID scores. The experiment results prove that our method is effective on different datasets and under various data settings. According to Friedman test, we think our improvements are significant.

\begin{table}[]
\centering
\caption{Comparison of Quality Scores (FID, KIDx${10}^{3}$) on the Cloud, USGS, HIT-UAV, SAR and Thermal datasets}
\renewcommand\arraystretch{1.5}
\begin{tabular}{lllll}
\hline
\toprule
\multicolumn{1}{c}{\multirow{2}{*}{\textbf{Methods}}} & \multicolumn{2}{c}{\textbf{StyleGAN2-ada}}                 & \multicolumn{2}{c}{\textbf{Ours}}                          \\ \cline{2-5} 
\multicolumn{1}{c}{}                         & \multicolumn{1}{c}{\textbf{FID}${\downarrow}$} & \multicolumn{1}{c}{\textbf{KID}${\downarrow}$} & \multicolumn{1}{c}{\textbf{FID}${\downarrow}$} & \multicolumn{1}{c}{\textbf{KID}${\downarrow}$} \\ 
\midrule
Cloud                                  & 33.34                   & 14.33                    & \textbf{30.76}                   & \textbf{12.07}                    \\
USGS                                 & 7.26                   & 2.29                     & \textbf{6.85}                    & \textbf{1.76}                    \\
HIT-UAV                                 & 83.77                   & 47.29                     & \textbf{81.57}                   & \textbf{43.22}                     \\
SAR                                & 12.93                   & 6.54                     & \textbf{11.81}                   & \textbf{5.12}                     \\
Thermal                               & 22.19                   & 9.86                     & \textbf{20.44}                   & \textbf{7.69}                     \\
\bottomrule
\end{tabular}
\label{table1-2}
\end{table}

\begin{table}[]
\centering
\caption{Comparison of Quality Scores (FID, KIDx${10}^{3}$) on the FFHQ256 and LSUN-cat datasets}
\renewcommand\arraystretch{1.5}
\begin{tabular}{lllll}
\hline
\toprule
\multicolumn{1}{c}{\multirow{2}{*}{\textbf{Methods}}} & \multicolumn{2}{c}{\textbf{StyleGAN2-ada}}                 & \multicolumn{2}{c}{\textbf{Ours}}                          \\ \cline{2-5} 
\multicolumn{1}{c}{}                         & \multicolumn{1}{c}{\textbf{FID}${\downarrow}$} & \multicolumn{1}{c}{\textbf{KID}${\downarrow}$} & \multicolumn{1}{c}{\textbf{FID}${\downarrow}$} & \multicolumn{1}{c}{\textbf{KID}${\downarrow}$} \\ 
\midrule
FFHQ256(2k)                                  & 26.41                   & 7.21                    & \textbf{24.20}                   & \textbf{5.44}                    \\
FFHQ256(10k)                                 & 10.62                   & 3.04                     & \textbf{9.38}                    & \textbf{2.38}                    \\
LSUN-cat(5k)                                 & 36.05                   & 10.44                     & \textbf{35.28}                   & \textbf{9.89}                     \\
LSUN-cat(30k)                                & 14.87                   & 5.31                     & \textbf{13.90}                   & \textbf{4.86}                     \\
\bottomrule
\end{tabular}
\label{table1-1}
\end{table}

\subsection{Generalizability Experiment}

To verify the generalizability of our method, we conduct experiments on RS images of different scales and modalities. 

First, we construct two new satellite datasets: Cloud, USGS, with lower resolution and higher resolution, respectively. Then, we use HIT-UAV dataset \citep{suo2023hit} to evaluate our method on infrared images. We collect HRSID-JPG \citep{wang2019sar} and SAR-Ship datasets \citep{wei2020hrsid} to evaluate our method on infrared images. For the thermal images, we couldn't find a suitable thermal satellite dataset. We conduct experiments on thermal natural image dataset for object classification \citep{ashfaq2021thermal}. The details of these datasets are available in the Appendix.

The experiment results are shown in Table \ref{table1-2}. Our method are effective on RS images of different scales and modalities. The visual results can be found in the Appendix.

In addition to this, we also perform our method on the natural image datasets: FFHQ256 and LSUN-cat under different data settings. The experiment results are shown in Table \ref{table1-1}, and our method still outperforms the baseline model. The generated images on these two datasets are available in the Appendix. These experiments prove the generalizability and effectiveness of our method.

\subsection{Ablation and Analysis}
\label{sec:abla}
In this section, we provide further ablation and analysis over different components of our method. 
\begin{small}
\begin{table}[]

\centering
\caption{Ablation study on the regularization components}
\renewcommand\arraystretch{1.5}
\resizebox{\linewidth}{15mm}{
\begin{tabular}{lllllll}
\hline
\toprule
\multicolumn{1}{c}{}                                   & \multicolumn{3}{c}{\textbf{NWPU}}                                                                                                   & \multicolumn{3}{c}{\textbf{PN}}                                                                              \\ \cline{2-7} 
\multicolumn{1}{c}{\multirow{-2}{*}{\textbf{Methods}}} & \multicolumn{1}{c}{\textbf{full}}  & \multicolumn{1}{c}{\textbf{300s}}  & \multicolumn{1}{c}{{ \textbf{15c}}} & \multicolumn{1}{c}{\textbf{full}}  & \multicolumn{1}{c}{\textbf{300s}}  & \multicolumn{1}{c}{\textbf{18c}} \\ 
\midrule
\textbf{Baseline}                                       & \multicolumn{1}{l}{11.97}          & \multicolumn{1}{l}{20.63}          & 19.64                                                    & \multicolumn{1}{l}{33.53}          & \multicolumn{1}{l}{44.21}          & 28.88                            \\ 
\textbf{Only UR}                                        & \multicolumn{1}{l}{11.28}          & \multicolumn{1}{l}{17.86}          & 18.26                                                    & \multicolumn{1}{l}{\textbf{31.48}} & \multicolumn{1}{l}{42.03}          & 27.67                            \\ 
\textbf{Only ER}                                        & \multicolumn{1}{l}{11.20}          & \multicolumn{1}{l}{18.24}          & 18.04                                                    & \multicolumn{1}{l}{32.87}          & \multicolumn{1}{l}{43.25}          & 26.61                            \\ 
\textbf{Ours}                                   & \multicolumn{1}{l}{\textbf{10.74}} & \multicolumn{1}{l}{\textbf{16.89}} & \textbf{17.88}                                           & \multicolumn{1}{l}{32.55}          & \multicolumn{1}{l}{\textbf{41.73}} & \textbf{25.22}                   \\ 
\bottomrule
\end{tabular}}
\label{table2}
\end{table}
\end{small}

\textbf{Regularization components}. We provide an ablation study containing four different variants. Table \ref{table2} presents the results of the ablation study on NWPU-RESISC45 and PatternNet datasets. The No Regularization version yields poor results as expected. Adding the uniformity regularization already brings significant improvement to the model under different data setups. The entropy regularization is also effective in most cases. As proposed in our final method, adding both uniformity and entropy regularization in the loss function of the generator achieves the best results. 

\begin{table}[]
\centering
\caption{Ablation study on temperature parameter ${\gamma}$}
\renewcommand\arraystretch{1.5}
\begin{tabular}{lcccc}
\toprule
${\gamma}$            & 2     & 3     & 5     & 7                            \\ 
\midrule
\textbf{FID} & \textbf{15.05} & 16.70 & 17.85 &  24.96 \\ 
\bottomrule
\end{tabular}
\label{table3}
\end{table}

\textbf{Temperature parameter and other hyperparameters}. First, we conduct the ablation study on temperature parameter ${\gamma}$ using StyleGAN2+ADA on NWPU-RESISC45 dataset with twenty classes. The FID scores are shown in Table \ref{table3}. Based on the experiment results, we set ${\gamma}=2$ in the following experiments. Then we conduct sensitive studies on hyperparameters using StyleGAN2+ADA on NWPU-RESISC45 and PatternNet dataset. The results are shown in Table \ref{table4} and Table \ref{table5}. The model gets better results when using features from both the generator and the discriminator, which is our final loss function. We set ${\lambda}_G=0.5$, ${\lambda}_D=0.5$, ${\delta}_G=0.1$, and ${\delta}_D=0.1$ for NWPU-RESISC45, AID, FFHQ256 and LSUN-cat datasets, ${\lambda}_G=0.2$, ${\lambda}_D=0.3$, ${\delta}_G=0.2$, and ${\delta}_D=0.2$ for PatternNet dataset in our experiments.

\begin{table}[]\small%
\centering
\caption{Ablation study on hyperparameters ${\lambda}_G$ and ${\lambda}_D$}
\renewcommand\arraystretch{1.5}
\resizebox{\linewidth}{15mm}{
\begin{tabular}{lcccccc}
\hline
\toprule
\multicolumn{1}{c}{}                                   & \multicolumn{3}{c}{\textbf{NWPU}}                                                                              & \multicolumn{3}{c}{\textbf{PN}}                                                            \\ \cline{2-7} 
\multicolumn{1}{c}{\multirow{-2}{*}{(\textbf{${\lambda}_G$, ${\lambda}_D$})}} & \multicolumn{1}{c}{\textbf{full}}  & \multicolumn{1}{c}{\textbf{300s}}  & { \textbf{15c}} & \multicolumn{1}{c}{\textbf{full}}  & \multicolumn{1}{c}{\textbf{300s}}  & \textbf{18c}   \\ 
\midrule
\textbf{(0, 0.2)}                                       & \multicolumn{1}{c}{11.24}          & \multicolumn{1}{c}{19.76}          & 18.33                               & \multicolumn{1}{c}{33.70}          & \multicolumn{1}{c}{42.48}          & 27.39          \\
\textbf{(0.2, 0)}                                       & \multicolumn{1}{c}{14.78}          & \multicolumn{1}{c}{20.98}          & 19.27                               & \multicolumn{1}{c}{34.63}          & \multicolumn{1}{c}{43.5}           & 29.24          \\ 
\textbf{(0.2,0.2)}                                      & \multicolumn{1}{c}{11.96}          & \multicolumn{1}{c}{17.24}          & 18.09                               & \multicolumn{1}{c}{\textbf{31.48}} & \multicolumn{1}{c}{41.73}          & \textbf{25.22} \\ 
\textbf{(0.5,0.5)}                                      & \multicolumn{1}{c}{\textbf{10.74}} & \multicolumn{1}{c}{\textbf{16.89}} & \textbf{17.88}                      & \multicolumn{1}{c}{33.24}          & \multicolumn{1}{c}{\textbf{41.39}} & 27.01          \\ 
\bottomrule
\end{tabular}}
\label{table4}
\end{table}

\begin{table}[]\small%
\centering
\caption{Ablation study on hyperparameters ${\delta}_G$ and ${\delta}_D$}
\renewcommand\arraystretch{1.5}
\resizebox{\linewidth}{15mm}{
\begin{tabular}{lcccccc}
\hline
\toprule
\multicolumn{1}{c}{}                                   & \multicolumn{3}{c}{\textbf{NWPU}}                                                                              & \multicolumn{3}{c}{\textbf{PN}}                                                            \\ \cline{2-7} 
\multicolumn{1}{c}{\multirow{-2}{*}{(\textbf{${\delta}_G$, ${\delta}_D$})}} & \multicolumn{1}{c}{\textbf{full}}  & \multicolumn{1}{c}{\textbf{300s}}  & {\textbf{15c}} & \multicolumn{1}{c}{\textbf{full}}  & \multicolumn{1}{c}{\textbf{300s}}  & \textbf{18c}   \\ 
\midrule
\textbf{(0, 0.1)}                                       & \multicolumn{1}{c}{12.08}          & \multicolumn{1}{c}{18.21}          & 19.03                               & \multicolumn{1}{c}{33.15}          & \multicolumn{1}{c}{42.18}          & 27.41          \\ 
\textbf{(0.1, 0)}                                       & \multicolumn{1}{c}{13.62}          & \multicolumn{1}{c}{21.14}          & 19.79                               & \multicolumn{1}{c}{35.63}          & \multicolumn{1}{c}{45.09}          & 28.56          \\
\textbf{(0.1,0.1)}                                      & \multicolumn{1}{c}{\textbf{10.74}} & \multicolumn{1}{c}{\textbf{16.89}} & \textbf{17.88}                      & \multicolumn{1}{c}{33.21}          & \multicolumn{1}{c}{42.53}          & 26.44          \\
\textbf{(0.2,0.2)}                                      & \multicolumn{1}{c}{11.29}          & \multicolumn{1}{c}{17.32} & 18.46                               & \multicolumn{1}{c}{\textbf{31.48}} & \multicolumn{1}{c}{\textbf{41.39}} & \textbf{25.22} \\ 
\bottomrule
\end{tabular}}
\label{table5}
\end{table}

\textbf{Feature selection}. Next, we conduct ablation studies on features from different network layers. As we mentioned before, different network layers are related to different levels of details in the generated image, and the earlier blocks of the network impact the coarse structure or content of the image. We conduct experiments on StyleGAN2+ADA model, which consists of 14 blocks. We choose 5 blocks for comparison, and the results are shown in Table \ref{table6}. We empirically choose the outputs of 8th block as the features in our method, which is also in line with our previous analysis.

\begin{table}[]
\centering
\caption{Ablation study on features from different blocks}
\renewcommand\arraystretch{1.5}
\begin{tabular}{lccccc}
\hline
\toprule
\textbf{Block} & 4    & 6    & 8    & 10  
  & 12    \\
\midrule
\textbf{FID}  & 17.26  & 15.92 & \textbf{15.05}  &  20.23  &  32.19  \\
\bottomrule
\end{tabular}
\label{table6}
\end{table}

\section{Conclusion}

In this work, we first investigate the properties of GAN models on RS images and observe that the training data size has a more significant impact on the GAN model's performance for RS images than for natural images. With fewer classes or samples per class, the distribution of generated samples in the feature space becomes clustered, and the features of individual samples get sparser, indicating lower information entropy. Based on this discovery, we conjecture that the amount of feature information learned by the GAN decreases dramatically as the training dataset reduces, leading to poor generation quality. Furthermore, we develope a structural causal model (SCM) to represent the data generation process and prove that the quality of generated images is correlated with the amount of feature information. Increasing the entropy of these features could improve the quality of generated images. Then, we design distribution-wise and sample-wise regularization schemes to increase the information entropy and enrich the feature information contained in the GAN model. Extensive experiments on eight RS datasets and three natural datasets show the advantages and versatility of our methods. This paper only focus on the GAN model, and our future work will involve analyzing the feature information learned by the diffusion model.

  %
  \section*{Data Availability}

  The benchmark datasets can be downloaded from the literature cited in Subsubsection \ref{sec:data}.

  \section*{Conflict of interest}
  
  The authors declare no conflict of interest.
  

  \bibliographystyle{spbasic}      
  \bibliography{egbib.bib}   
  
\clearpage
\appendix
\section{Proofs and Theoretical Analysis}

\subsection{Proof of \textit{Theorem} 1}

The global minimum of $\mathcal{L}_{\text {AlignMax}}$ is achieved when the first term (distance between real and generated distributions) is minimized (i.e., equal to zero) and the second term (entropy) is maximized. Commonly, the uniform distribution on $(0, 1)^{d_c}$ gets the maximum entropy.

\textbf{Step 1}. First, we show that there exists a smooth function $\mathbf{g}_{\ast}: \mathcal{X} \rightarrow (0, 1)^{d_c}$ which gets the global minimum of $\mathcal{L}_{\text {AlignMax}}$. Consider the function $\mathbf{f}^{-1}_{1:{d_c}}: \mathcal{X}\rightarrow \mathcal{C}$, i.e., the inverse of the true mixing $\mathbf{f}$, restricted to its first $d_c$ dimensions. Further, we have $\mathbf{f}^{-1}(x)_{1:{d_c}} = \mathbf{c}$ by definition.

Then we construct a function $\mathbf{k}:\mathcal{C} \rightarrow (0, 1)^{d_c}$ which maps $\mathbf{c}$ to a uniform random variable on $(0, 1)^{d_c}$ using a recursive method known as the Darmois construction.

Specifically, we define
\begin{equation}
\begin{aligned}
k_{i}(\mathbf{c}): & = & F_{i}\left(c_{i} \mid \mathbf{c}_{1: i-1}\right) & = & \mathbb{P}\left(C_{i} \leq c_{i} \mid \mathbf{c}_{1: i-1}\right), \nonumber
\end{aligned}
\end{equation}
where $i= 1, \ldots, d_{c}$, $F_i$ represents the conditional cumulative distribution function (CDF) of $c_i$ given $\mathbf{c}_{1: i-1}$. By construction, $\mathbf{k}(\mathbf{c})$ is uniformly distributed on $(0, 1)^{d_c}$. Moreover, $\mathbf{k}$ is smooth by the assumption that $p_\mathbf{z}$ (and hence $p_\mathbf{c}$) is a smooth density.
Finally, we define
\begin{equation}
\begin{aligned}
\mathbf{g}^{*}: & = & \mathbf{k} \circ \mathbf{f}_{1: d_{c}}^{-1}: \mathcal{X} \rightarrow(0,1)^{d_{c}} \nonumber
\end{aligned}
\end{equation}
which is a smooth function since it is a composition of two smooth functions.

Next, we proof $\mathbf{g}^{*}$ gets the global minimum of $\mathcal{L}_{\text {AlignMax}}$.
Using $\mathbf{f}^{-1}(\mathbf{x})_{1:{d_c}} = \mathbf{c}$ and $\mathbf{f}^{-1}(\widetilde{\mathbf{x}})_{1:{d_c}} = \widetilde{\mathbf{c}}$, we have
\begin{scriptsize}
\begin{equation}
\begin{aligned}
\mathcal{L}_{\text {AlignMax}}\left(\mathbf{g}^{*}\right) & =\mathbb{E}_{(\mathbf{x}, \tilde{\mathbf{x}}) \sim p_{(\mathbf{x}, \tilde{\mathbf{x}})}}\left[\left\|\mathbf{g}^{*}(\mathbf{x})-\mathbf{g}^{*}(\tilde{\mathbf{x}})\right\|_{2}^{2}\right]-H\left(\mathbf{g}^{*}(\mathbf{x})\right) \\
& =\mathbb{E}_{(\mathbf{x}, \tilde{\mathbf{x}}) \sim p_{(\mathbf{x}, \tilde{\mathbf{x}})}}\left[\|\mathbf{k}(\mathbf{c})-\mathbf{k}(\tilde{\mathbf{c}})\|_{2}\right]-H(\mathbf{k}(\mathbf{c})) \\
& =0 \nonumber
\end{aligned}
\end{equation}
\end{scriptsize}

where in the last step we utilize the fact that $\mathbf{c} = \widetilde{\mathbf{c}}$ almost surely with respect to the ground truth generative process described in Fig.\ref{fig2-2}. As a result, the first term is zero. Moreover, since $\mathbf{k}(\mathbf{c})$ is uniformly distributed on $(0, 1)^{d_c}$ and the uniform distribution on the unit hypercube has zero entropy, the second term is also zero.

Next, let $\mathbf{g}:\mathcal{X} \rightarrow (0, 1)^{d_c}$ be any smooth function which attains the global minimum of of $\mathcal{L}_{\text {AlignMax}}$, i.e.,
\begin{scriptsize}
\begin{equation}
\begin{aligned}
\mathcal{L}_{\mathrm{AlignMax}}(\mathbf{g}) & = \mathbb{E}_{(\mathbf{x}, \tilde{\mathbf{x}}) \sim p_{(\mathbf{x}, \tilde{\mathbf{x}})}}\left[\|\mathbf{g}(\mathbf{x})-\mathbf{g}(\tilde{\mathbf{x}})\|_{2}^{2}\right]-H(\mathbf{g}(\mathbf{x})) & = 0 \nonumber
\end{aligned}
\end{equation}
\end{scriptsize}

Define $\mathbf{h}: = \mathbf{g} \circ \mathbf{f}: \mathcal{Z} \rightarrow(0,1)^{d_{c}}$ which is smooth because both $g$ and $f$ are smooth.
Given $\mathbf{x}=\mathbf{f}(\mathbf{z})$, we can get

\begin{equation}
\begin{aligned}
\mathbb{E}_{(\mathbf{x}, \tilde{\mathbf{x}}) \sim p_{(\mathbf{x}, \tilde{\mathbf{x}})}}\left[\|\mathbf{h}(\mathbf{z})-\mathbf{h}(\tilde{\mathbf{z}})\|_{2}\right] & =0, \\
H(\mathbf{h}(\mathbf{z})) & =0 .
\end{aligned}
\label{equation9}
\end{equation}
The first equation suggests that $\hat{\mathbf{c}}=\mathbf{h}(\mathbf{z})_{1: d_{c}}= \mathbf{h}(\tilde{\mathbf{z}})_{1: d_{c}}$ must hold (almost surely with respect to $p$), and the second equation suggests that $\hat{\mathbf{c}}=\mathbf{h}(\mathbf{z})$ must be uniformly distributed on $(0,1)^{d_c}$.

\textbf{Step 2}. Next, we show that $\mathbf{h}(\mathbf{z}) = \mathbf{h}(\mathbf{c}, \mathbf{\epsilon})$ can only depend on the true content $\mathbf{c}$ and not on any of the noise variables $\mathbf{\epsilon}$.

Suppose for a contradiction that $\mathbf{h}_{c}(\mathbf{c}, \mathbf{\epsilon}):=\mathbf{h}(\mathbf{c}, \mathbf{\epsilon})_{1: d_{c}}=\mathbf{h}(\mathbf{z})_{1: d_{c}}$ depends on some component of the noise variable $\mathbf{\epsilon}$:

\begin{equation}
\begin{aligned}
\exists l \in\left\{1, \ldots, d_{\epsilon}\right\},\left(\mathbf{c}^{*}, \mathbf{\epsilon}^{*}\right) \in \mathcal{C} \times \mathcal{E}, \text { s.t. } \frac{\partial \mathbf{h}_{c}}{\partial {\epsilon}_{l}}\left(\mathbf{c}^{*}, \mathbf{\epsilon}^{*}\right) \neq 0 
\end{aligned}
\label{equation10}
\end{equation}

that is, we assume that the partial derivative of $\mathbf{h}_c$ with respect to some noise variable ${\epsilon}_l$ is non-zero at some point $\mathbf{z}^{*}=\left(\mathbf{c}^{*}, \mathbf{\epsilon}^{*}\right) \in \mathcal{Z}=\mathcal{C} \times \mathcal{E}$.

Since $\mathbf{h}$ is smooth, so is $\mathbf{h}_c$. Therefore, $\mathbf{h}_c$ has continuous (first) partial derivatives. By continuity of the partial derivative, $\frac{\partial \mathbf{h}_{c}}{\partial {\epsilon}_{l}}$ must be non-zero in a neighbourhood of $(\mathbf{c}^{*},\mathbf{\epsilon}^{*})$, i.e.
\begin{equation}
\begin{aligned}
\exists \eta>0 \text { s.t. } {\epsilon}_{l} \mapsto \mathbf{h}_{c}\left(\mathbf{\epsilon}^{*},\left(\mathbf{\epsilon}_{-l}^{*}, {\epsilon}_{l}\right)\right) \text { is strictly monotonic on }\\
\left({\epsilon}_{l}^{*}-\eta, {\epsilon}_{l}^{*}+\eta\right),
\end{aligned}
\label{equation11}
\end{equation}
where $\mathbf{\epsilon}_{-l} \in \mathcal{E}_{-l}$ denotes the vector of remaining noise variables except ${\epsilon}_l$.

Next, define the auxiliary function $\psi: \mathcal{C} \times \mathcal{E} \times \mathcal{E} \rightarrow \mathbb{R}_{\geq 0}$ as follows:
\begin{equation}
\begin{aligned}
\psi(\mathbf{c}, \mathbf{\epsilon}, \tilde{\mathbf{\epsilon}}): & = & \left|\mathbf{h}_{c}(\mathbf{c}, \mathbf{\epsilon})-\mathbf{h}_{c}(\mathbf{c}, \tilde{\mathbf{\epsilon}})\right| \geq 0 \nonumber
\end{aligned}
\end{equation}

To obtain a contradiction to the invariance condition under assumption Eq.\ref{equation10}, it remains to show that $\psi$ is strictly positive with probability greater than zero (w.r.t. $p$).

First, the strict monotonicity from Eq.\ref{equation11} that suggests
\begin{equation}
\begin{aligned}
\psi\left(\mathbf{c}^{*},\left(\mathbf{\epsilon}_{-l}^{*}, {\epsilon}_{l}\right),\left(\mathbf{\epsilon}_{-l}^{*}, \tilde{\epsilon}_{l}\right)\right)>0, \\
\forall\left({\epsilon}_{l}, \tilde{\epsilon}_{l}\right) \in\left({\epsilon}_{l}^{*}, {\epsilon}_{l}^{*}+\eta\right) \times\left({\epsilon}_{l}^{*}-\eta, {\epsilon}_{l}^{*}\right)
\end{aligned}
\label{equation13}
\end{equation}

Note that in order to obtain the strict inequality in Eq.\ref{equation13}, it is important that ${\epsilon}_l$ and $\widetilde{\epsilon}_l$ take values in disjoint open subsets of the interval $({\epsilon}_{l}^{*}-\eta, {\epsilon}_{l}^{*}+\eta)$ from Eq.\ref{equation11}.

Since $\psi$ is a composition of continuous functions (absolute value of the difference of two continuous functions), it is also continuous. Considering the open set $\mathbb{R}_{>0}$, it is known that pre-images (or inverse images) of open sets under a continuous function are always open. For the continuous function $\psi$, this pre-image corresponds to an open set
\begin{equation}
\begin{aligned}
\mathcal{U} \subseteq \mathcal{C} \times \mathcal{E} \times \mathcal{E} \nonumber
\end{aligned}
\end{equation}
in the domain of $\psi$ on which $\psi$ is strictly positive.

Moreover, due to Eq.\ref{equation13}
\begin{equation}
\begin{aligned}
\left\{\mathbf{c}^{*}\right\} \times\left(\left\{\mathbf{\epsilon}_{-l}^{*}\right\} \times\left({\epsilon}_{l}^{*}, {\epsilon}_{l}^{*}+\eta\right)\right) \times\left(\left\{\mathbf{\epsilon}_{-l}^{*}\right\} \times\left({\epsilon}_{l}^{*}-\eta, {\epsilon}_{l}^{*}\right)\right) \subset \mathcal{U} 
\end{aligned}
\label{equation15}
\end{equation}
so $\mathcal{U}$ is non-empty.

Next, by assumption (iii), there exists at least one subset $A \subseteq {1, ..., d_{\epsilon}}$ of changing noise variables such that $l in A$ and $p_A(A) > 0$; pick one such subset and call it $A$.

Then, also by assumption (iii), for any $\mathbf{\epsilon}_A \in \mathcal{E}_A$, there is an open subset $\mathcal{O}(\mathbf{\epsilon}_A) \subseteq \mathcal{E}_A$ containing $\mathbf{\epsilon}_A$, such that $p_{\tilde{\mathbf{\epsilon}}_{A} \mid \mathbf{\epsilon}_{A}}\left(\cdot \mid \mathbf{\epsilon}_{A}\right)>0$ within $\mathcal{O}\left(\mathbf{\epsilon}_{A}\right)$.

Define the following space
\begin{equation}
\begin{aligned}
\mathcal{R}_{A}: & = & \left\{\left(\mathbf{\epsilon}_{A}, \tilde{\mathbf{\epsilon}}_{A}\right): \mathbf{\epsilon}_{A} \in \mathcal{E}_{A}, \tilde{\mathbf{\epsilon}}_{A} \in \mathcal{O}\left(\mathbf{\epsilon}_{A}\right)\right\} \nonumber
\end{aligned}
\end{equation}
and, recalling that  $A^{\mathrm{c}}=\left\{1, \ldots, d_{\epsilon}\right\} \backslash A$ denotes the complement of $A$, define
\begin{equation}
\begin{aligned}
\mathcal{R}:=\mathcal{C} \times 
\mathcal{E}_{A^{\mathrm{c}}} \times \mathcal{R}_{A} \nonumber
\end{aligned}
\end{equation}
which is a topological subspace of $\mathcal{C} \times \mathcal{E} \times \mathcal{E}$.

By assumptions (ii) and (iii), $p_\mathbf{z}$ is smooth and fully supported, and $p_{\tilde{\mathbf{\epsilon}}_{A} \mid \mathbf{\epsilon}_{A}}\left(\cdot \mid \mathbf{\epsilon}_{A}\right)$ is smooth and fully supported on $\mathcal{O}(\mathbf{\epsilon}_A)$ for any $\mathbf{\epsilon}_A \in \mathcal{E}_A$. Therefore, the measure $\mu_{\left(\mathbf{c}, \mathbf{\epsilon}_{A^{\mathrm{c}}}, \mathbf{\epsilon}_{A}, \tilde{\mathbf{\epsilon}}_{A}\right) \mid A}$ has fully supported, strictly-positive density on $\mathcal{R}$ w.r.t. a strictly positive measure on $\mathcal{R}$. In other words, $p_{z} \times p_{\tilde{\epsilon}_{A} \mid {\epsilon}_{A}}$ is fully supported (i.e., strictly positive) on $\mathcal{R}$.

Now consider the intersection $\mathcal{U}\cap \mathcal{R}$ of the open set $\mathcal{U}$ with the topological subspace $\mathcal{R}$. Since $\mathcal{U}$ is open, by definition of topological subspaces, the intersection $\mathcal{U}\cap \mathcal{R} \subseteq \mathcal{R}$ is open in $\mathcal{R}$ and thus has the same dimension as $\mathcal{R}$ if non-empty.

Moreover, since $\mathcal{O}\left(\mathbf{\epsilon}_{A}^{*}\right)$ is open containing  $\mathbf{\epsilon}_{A}^{*}$, there exists $\eta^{\prime}>0$ such that  $\left\{\mathbf{\epsilon}_{-l}^{*}\right\} \times\left({\epsilon}_{l}^{*}-\eta^{\prime}, {\epsilon}_{l}^{*}\right) \subset   \mathcal{O}\left(\mathbf{\epsilon}_{A}^{*}\right)$. Thus, for $\eta^{\prime \prime}=\min \left(\eta, \eta^{\prime}\right)>0$,
\begin{equation}
\begin{aligned}
\left\{\mathbf{c}^{*}\right\} \times\left\{\mathbf{\epsilon}_{A^{c}}^{*}\right\} \times\left(\left\{\mathbf{\epsilon}_{A \backslash\{l\}}^{*}\right\} \times\left({\epsilon}_{l}^{*}, {\epsilon}_{l}^{*}+\eta\right)\right) \\
\times\left(\left\{\mathbf{\epsilon}_{A \backslash\{l\}}^{*}\right\} \times\left({\epsilon}_{l}^{*}-\eta^{\prime \prime}, {\epsilon}_{l}^{*}\right)\right) \subset \mathcal{R} . \nonumber
\end{aligned}
\end{equation}
In particular, this implies that
\begin{equation}
\begin{aligned}
\left\{\mathbf{c}^{*}\right\} \times\left(\left\{\mathbf{\epsilon}_{-l}^{*}\right\} \times\left({\epsilon}_{l}^{*}, {\epsilon}_{l}^{*}+\eta\right)\right) \times\left(\left\{\mathbf{\epsilon}_{-l}^{*}\right\} \times\left({\epsilon}_{l}^{*}-\eta^{\prime \prime}, {\epsilon}_{l}^{*}\right)\right) \subset \mathcal{R}
\end{aligned}
\label{equation19}
\end{equation}
Now, since  $\eta^{\prime \prime} \leq \eta$ , the LHS of Eq.\ref{equation19} is also in $\mathcal{U}$, so the intersection $\mathcal{U} \cap \mathcal{R}$ is non-empty.

In summary, the intersection $\mathcal{U} \cap \mathcal{R} \subseteq \mathcal{R}$:
\begin{itemize}
    \item[$ \bullet $] is non-empty since both $\mathcal{U}$ and $\mathcal{R}$ contain the LHS of Eq.\ref{equation15};
    \item [$ \bullet $]is an open subset of the topological subspace $\mathcal{R}$ of $\mathcal{C} \times \mathcal{E} \times \mathcal{E}$ since it is the intersection of an open set, $\mathcal{U}$, with $\mathcal{R}$;
    \item[$ \bullet $] satisfies $\psi>0$ since this holds for all of $\mathcal{U}$;
    \item[$ \bullet $] is fully supported w.r.t. the generative process since this holds for all of  $\mathcal{R}$.
\end{itemize}

As a consequence,
\begin{equation}
\begin{aligned}
\mathbb{P}(\psi(\mathbf{c}, \mathbf{\epsilon}, \tilde{\mathbf{\epsilon}})>0 \mid A) \geq \mathbb{P}(\mathcal{U} \cap \mathcal{R})>0
\end{aligned}
\end{equation}
where $\mathbb{P}$ denotes probability w.r.t. the true generative process $p$. Since $p_A(A)>0$, this is a contradiction to the invariance in Eq.\ref{equation9}.

Hence, $\mathbf{h}_c(\mathbf{c},\mathbf{\epsilon})$ does not depend on any noise variable ${\epsilon}_l$. It is thus only a function of $\mathbf{c}$, i.e., $\hat{\mathbf{c}} = \mathbf{h}_c(\mathbf{c})$.

\textbf{Step 3}. Finally, we show that the mapping $\hat{\mathbf{c}} = \mathbf{h}(\mathbf{c})$ is invertible. We use the following result from \citep{zimmermann2021contrastive}.
\begin{proposition}
    Let $\mathcal{M}$, $\mathcal{N}$ be simply connected and oriented $\mathcal{C}^{1}$ manifolds without boundaries and $h: \mathcal{M} \rightarrow \mathcal{N}$ be a differentiable map. Further, let the random variable $\mathbf{z} \in \mathcal{M}$ be distributed according to $\mathbf{z} \sim p(\mathbf{z})$ for a regular density function $p$, i.e., $0<p<\infty$. If the pushforward $p_{\# h}(\mathbf{z})$ of $p$ through $h$ is also a regular density, i.e., $0<p_{\# h}<\infty$, then $h$ is a bijection.
\end{proposition}

We apply this result to the simply connected and oriented $\mathcal{C}^{1}$ manifolds without boundaries $\mathcal{M}=\mathcal{C}$ and $\mathcal{N}=(0,1)^{d_{c}}$, and the smooth (hence, differentiable) map $\mathbf{h}: \mathcal{C} \rightarrow(0,1)^{d_{c}}$ which maps the random variable $\mathbf{c}$ to a uniform random variable $\hat{\mathbf{c}}$  (as established in Eq.\ref{equation9}).

Since both $p_{\mathbf{c}}$ (by assumption) and the uniform distribution (the pushforward of $p_{\mathbf{c}}$ through $\mathbf{h}$) are regular densities in the sense of Proposition 1, we conclude that $\mathbf{h}$ is a bijection, i.e., invertible.

We have shown that for any smooth $\mathbf{g}: \mathcal{X} \rightarrow(0,1)^{d_{c}}$ which minimises $\mathcal{L}_{\text {AlignMax}}$, we have that  $\hat{\mathbf{c}}=\mathbf{g}(\mathbf{x})=\mathbf{h}(\mathbf{c})$ for a smooth and invertible $\mathbf{h}: \mathcal{C} \rightarrow(0,1)^{d_{c}}$, i.e., $\mathbf{c}$ is block-identified by $\mathbf{g}$.

\subsection{Proof of \textit{Theorem} 2 And \textit{Theorem} 3}

In this section, we present proofs for the \textit{Theorem} 2 and \textit{Theorem} 3 in main paper Sections 4.1. These two theorems illustrate the deep relations between the Gaussian kernel $K:S^d {\times} S^d\ {\rightarrow}\ \mathbb{R}$ and the uniform distribution on the unit hypersphere $S^d$. As we will show below, these properties directly follow well-known results on strictly positive definite kernels.
\begin{definition}
(Strict positive definiteness). A symmetric and lower semi-continuous kernel $K$ on $A {\times} A$ (where $A$ is infinite and compact) is called strictly positive definite if for every finite signed Borel measure $\mu$ supported on $A$ whose energy
\begin{eqnarray}
I_{K}[\mu] \triangleq \int_{\mathcal{S}^{d}} \int_{\mathcal{S}^{d}} K(u, v) \mathrm{d} \mu(v) \mathrm{d} \mu(u)
\end{eqnarray}
is well defined, we have $I_{K}[{\mu}]\ {\geq}\ 0$, where equality holds only if ${\mu}\ {\equiv}\ 0$ on the $\sigma$-algebra of Borel subsets of $A$.
\end{definition}

\begin{definition}
Let M($S^d$) be the set of Borel probability measures on $S^d$.
\end{definition}

\begin{definition}
(Uniform distribution on $S^d$). ${\sigma}^d$ denotes the normalized surface area measure on $S^d$.
\end{definition}
According to Borodachov et al.\citep{gaussianK}, we can get the following two results.
\begin{lemma}
(Strict positive definiteness of $G_{\gamma}$). For ${\gamma}\ >\ 0$, the Gaussian kernel is $G_{\gamma}(x, y)\ {\triangleq}\ e^{-\gamma\|x-y\|_{2}^{2}}$ strictly positive definite on $S^d {\times} S^d$.
\end{lemma}
\begin{lemma}
(Strictly positive definite kernels on $S^d$). Consider kernel $K_{f}: \mathcal{S}^{d} \times \mathcal{S}^{d} \rightarrow(-\infty,+\infty] $ of the form:
\begin{eqnarray}
K_{f}(u, v) \triangleq f\left(\|u-v\|_{2}^{2}\right)
\end{eqnarray}
If $K_f$ is strictly positive definite on $S^d {\times} S^d$ and $I_{K_f}[{\sigma}^d]$ is finite, then ${\sigma}^d$ is the unique measure (on Borel subsets of $S^d$) in the solution of $\min _{\mu \in \mathcal{M}\left(\mathcal{S}^{d}\right)} I_{K_{f}}[\mu]$, and the normalized counting measures associated with any $K_f$ -energy minimizing sequence of N-point configurations on $S^d$ converges weak* to ${\sigma}^d$. In particular, this conclusion holds whenever $f$ has the property that $-f'(t)$ is strictly completely monotone on $(0, 4]$ and $I_{K_f}[{\sigma}^d]$ is finite.
\end{lemma}
Based on \textbf{Lemma 1} and \textbf{Lemma 2}, we now can get two propositions.
\begin{proposition}
${\sigma}^d$ is the unique solution (on Borel subsets of $S_d$) of
\begin{eqnarray}
\min _{\mu \in \mathcal{M}\left(\mathcal{S}^{d}\right)} I_{G_{t}}[\mu]=\min _{\mu \in \mathcal{M}\left(\mathcal{S}^{d}\right)} \int_{\mathcal{S}^{d}} \int_{\mathcal{S}^{d}} G_{t}(u, v) \mathrm{d} \mu(v) \mathrm{d} \mu(u)
\end{eqnarray}
\end{proposition}

\begin{definition}
   ($\text{Weak}^{*}$ convergence of measures). A sequence of Borel measures  $\left\{\mu_{n}\right\}_{n=1}^{\infty}$ in $\mathbb{R}^{p}$ converges $\text{weak}^{*}$ to a Borel measure $\mu$ if for all continuous function  $f: \mathbb{R}^{p} \rightarrow \mathbb{R}$, we have
   \begin{eqnarray}
   \lim _{n \rightarrow \infty} \int f(x) \mathrm{d} \mu_{n}(x)=\int f(x) \mathrm{d} \mu(x)
   \end{eqnarray}
\end{definition}

\begin{proposition}
For each $N\ >\ 0$, the $N$ point minimizer of the average pairwise potential is
\begin{eqnarray}
\mathbf{u}_{N}^{*} & = & \underset{u_{1}, u_{2}, \ldots, u_{N} \in \mathcal{S}^{d}}{\arg \min } \sum_{1 \leq i<j \leq N} G_{t}\left(u_{i}, u_{j}\right)
\end{eqnarray}
The normalized counting measures associated with the $\left\{\mathbf{u}_{N}^{*}\right\}_{N=1}^{\infty}$ sequence converge $\text{weak}^{*}$ to $ \sigma_{d}$.
\end{proposition}
According to \textbf{Proposition 2} and \textbf{Proposition 3}, the \textit{Theorem} 2 and \textit{Theorem} 3 can be naturally proven.

\subsection{Information Theory}

In Section \ref{sec:intro}, we mentioned that sparser features contain less information, and the uniform distribution gets the maximum information entropy. This conclusion is inspired from \textbf{Theorem 2.6.4} and \textbf{Theorem 8.2.2} in \citep{information}.

\textbf{Theorem 2.6.4} $H (X) \le log |X|$, where $|X|$ denotes the number of elements in the range of $X$, with equality if and only $X$ has a uniform distribution over $X$.

\textbf{Theorem 8.2.2} The typical set  $A_{\epsilon}^{(n)}$  has the following properties:
\begin{itemize}
    \item $\operatorname{Pr}\left(A_{\epsilon}^{(n)}\right)>1-\epsilon$ for $n$ sufficiently large.
    \item $\operatorname{Vol}\left(A_{\epsilon}^{(n)}\right) \leq 2^{n(h(X)+\epsilon)}$ for all $n$.
    \item $\operatorname{Vol}\left(A_{\epsilon}^{(n)}\right) \geq(1-\epsilon) 2^{n(h(X)-\epsilon)}$ for  $n$ sufficiently large.  ''\\
\end{itemize}

\textbf{Theorem 2.6.4} proves that the uniform distribution gets the maximum information entropy, and \textbf{Theorem 8.2.2} proves that sparser features contain less information.

\section{Experiment Details}
\subsection{Datasets}

\textbf{Cloud Datset}: We collect three datasets: 38cloud \citep{cloud-1}, Sentinel-2 Cloud Mask \citep{aybar2022cloudsen12} and SPCRAS \citep{hughes2014automated}. The 38cloud dataset consists of images collected from the Landsat 8 satellite. The size of images varies from 7000x7000 to 8000x8000 pixels. The spatial resolution of this dataset is 30m. The Sentinel-2 Cloud Mask dataset consists of images collected from the Sentinel-2 satellite. Each image is 1022x1022 pixels in size. The spatial resolution of this dataset is 20m. The SPCRAS dataset consists of images collected from the Landsat 8 satellite. Each image is 1000x1000 pixels in size. The spatial resolution of this dataset is 30m. These datasets are filtered, cropped and downsampled to construct a new dataset: 'Cloud'. The Cloud dataset has 1,386 images in total, and each image has 128x128 pixels and kilometer resolution.

\textbf{USGS Datset}: We collect the USGS image dataset of SIRI-WHU \citep{zhao2015dirichlet}. The dataset was acquired from United States Geological Survey (USGS) covering Montgomery, Ohio, USA. It has four classes: farms, forests, parking lot and residential area. It consists of one image of 10000x9000 pixels in size. The spatial resolution of this image is 2 feet. We first crop image patches of 64x64 pixels from the USGS image. Then, these patches are upsampled to 256x256 pixels. In this way, we construct a new dataset which has 21,840 images in total, and almost sub-centimeter resolution.

\textbf{HIT-UAV Dataset}: The HIT-UAV dataset \citep{suo2023hit} comprises 2,898 infrared thermal images extracted from 43,470 frames in hundreds of videos captured by Unmanned Aerial Vehicles (UAVs) in various scenarios. Each image is 256x256 pixels in size, and only has one channel.

\textbf{SAR Dataset}: We collect HRSID-JPG \citep{wang2019sar} and SAR-Ship datasets \citep{wei2020hrsid}. The HRSID-JPG dataset comprises 5,604 high-resolution SAR images and 16,951 ship instances. The SAR-Ship-Dataset has 43,819 ship slices. We filter these datasets to remove images with too much background. Then we constuct a new Radar dataset which has 9,407 images in total, and each image is 256x256 pixels in size and only has one channel.

\textbf{Thermal Dataset}: We conduct experiments on thermal natural image dataset for object classification \citep{ashfaq2021thermal}. The images were captured using Seek Thermal and FLIR. There are three classes: cat, car and man. This dataset was filtered, center-cropped and resized. We finally get a new dataset which has 5,543 images in total, and each image is 256x256 pixels in size.

\subsection{Implementation Details}

For the infrared, thermal, and Radar images, their image shapes are similar to the natural images, so we do not need to adjust the network architecture. During our experiments, we made some adjustments to the hyperparameters of the loss function as shown in Table \ref{table_a}.

\begin{table}[]
\centering
\caption{The choice of hyperparameters ${\lambda}_G$, ${\lambda}_D$, ${\delta}_G$ and ${\delta}_D$.}
\renewcommand\arraystretch{1.5}
\begin{tabular}{ccccc}
\hline
\toprule
\textbf{Hyperparameters} & ${\lambda}_G$ & ${\lambda}_D$ & ${\delta}_G$ & ${\delta}_D$ \\ \midrule
Cloud                    & 0.5         & 0.5         & 0.2         & 0.2         \\
USGS                     & 0.3         & 0.3         & 0.2         & 0.2         \\
HIT-UAV                  & 0.2         & 0.2         & 0.1         & 0.1         \\
SAR                      & 0.2         & 0.2         & 0.1         & 0.1         \\
Thermal                  & 0.5         & 0.5         & 0.2         & 0.2         \\ \bottomrule
\end{tabular}
\label{table_a}
\end{table}

\section{Visual Results}
\begin{figure*}[h]
\centering
\includegraphics[width=1\textwidth]{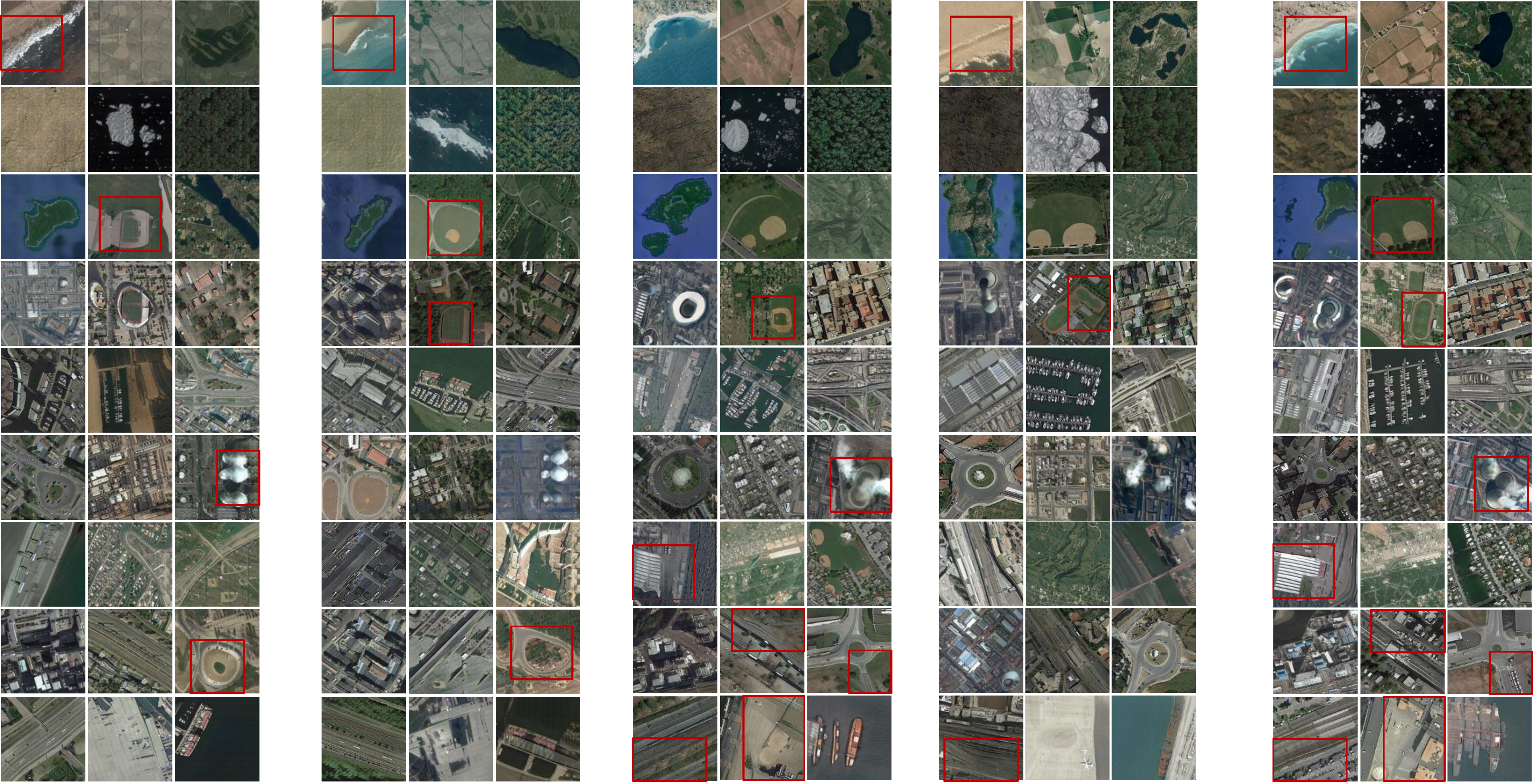}
\caption{The RS images generated by BigGAN+ADA(first from the left), BigGAN+ADA+Lecam(second from the left), StyleGAN2+ADA(middle), StyleGAN2+ADA+Lecam(second from the right) and our method(first from the right) trained on NWPU dataset.}
\label{fig6}
\end{figure*}

\begin{figure*}[htbp]
\centering
\includegraphics[width=1\textwidth]{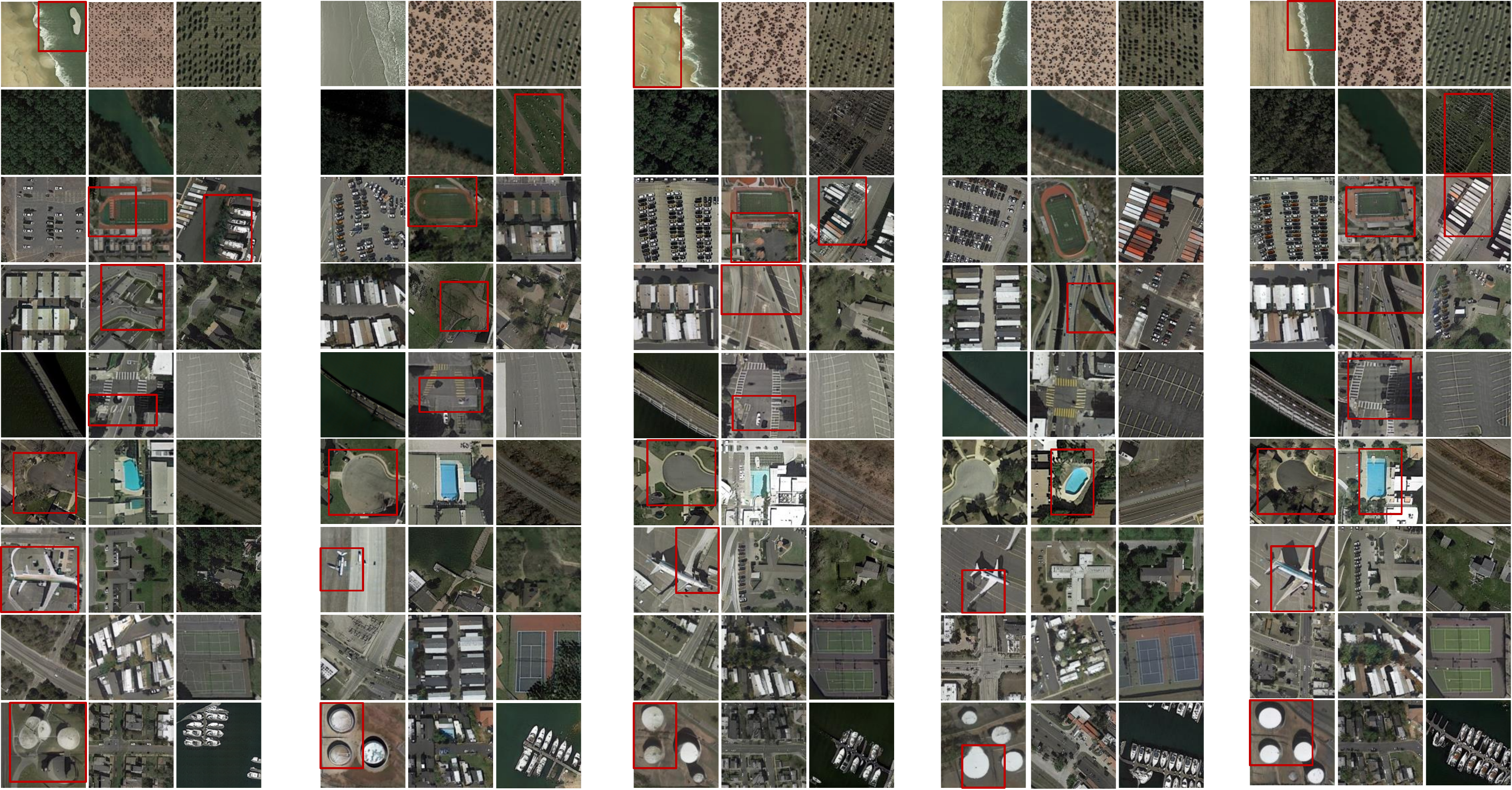}
\caption{The RS images generated by BigGAN+ADA(first from the left), BigGAN+ADA+Lecam(second from the left), StyleGAN2+ADA(middle), StyleGAN2+ADA+Lecam(second from the right) and our method(first from the right) trained on PN dataset.}
\label{fig7}
\end{figure*}

\begin{figure*}[htbp]
\centering
\includegraphics[width=1\textwidth]{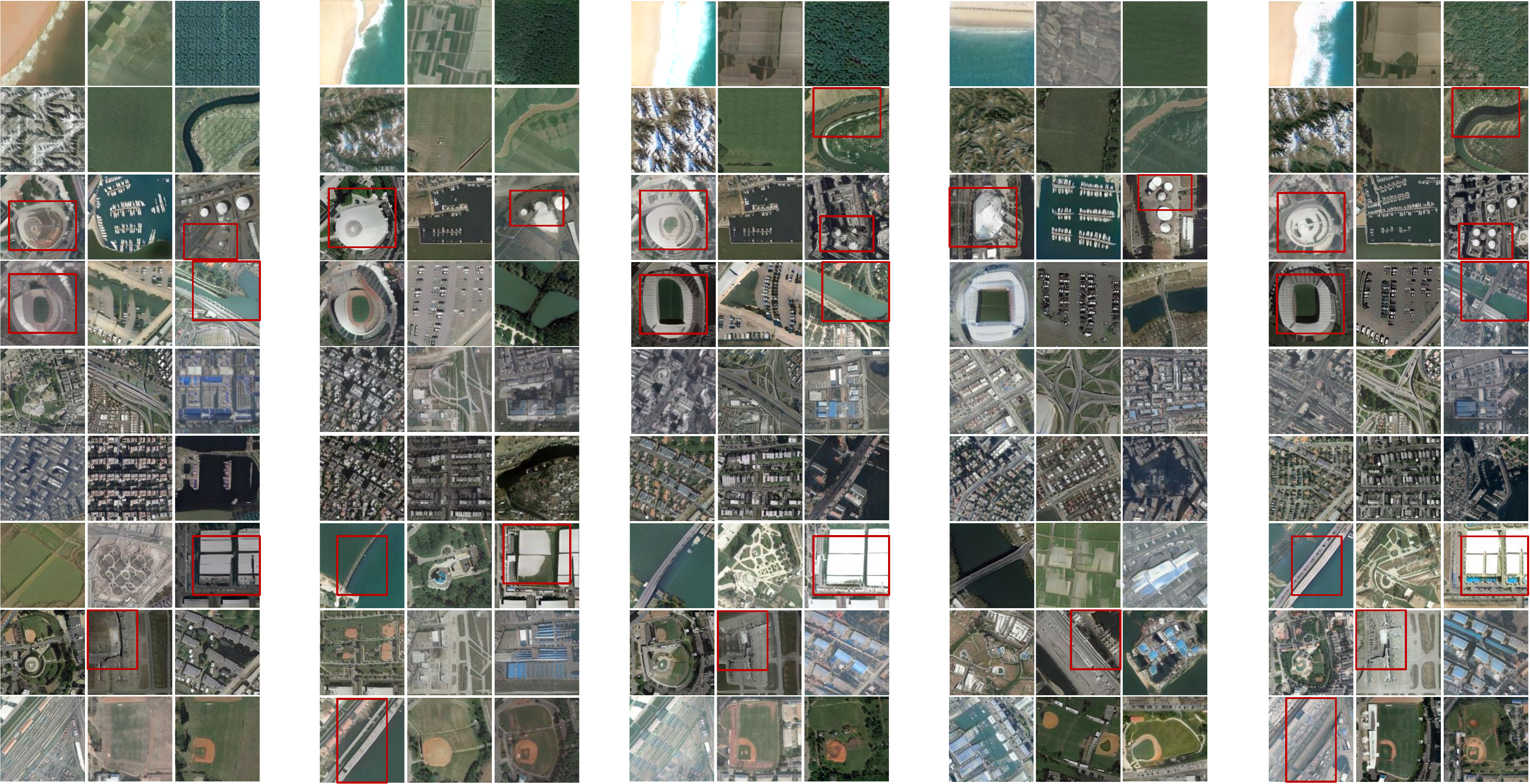}
\caption{The RS images generated by BigGAN+ADA(first from the left), BigGAN+ADA+Lecam(second from the left), StyleGAN2+ADA(middle), StyleGAN2+ADA+Lecam(second from the right) and our method(first from the right) trained on AID dataset.}
\label{fig8}
\end{figure*}

\begin{figure*}[htpb]
    \centering
    \subfigure[]{\includegraphics[width=0.25\textwidth]{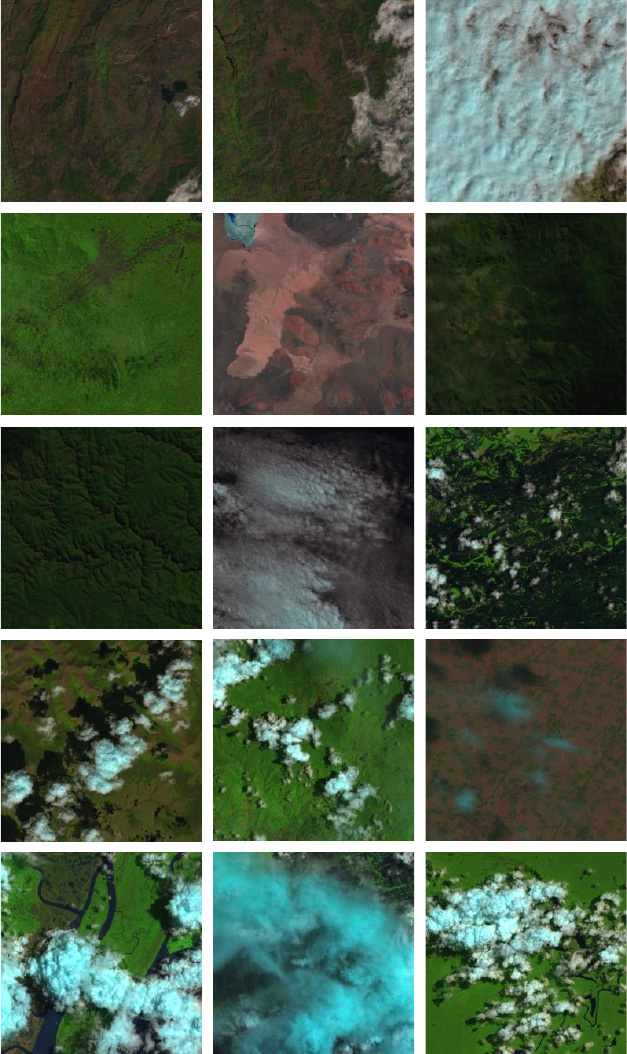}} \hspace{5pt}
    \subfigure[]{\includegraphics[width=0.25\textwidth]{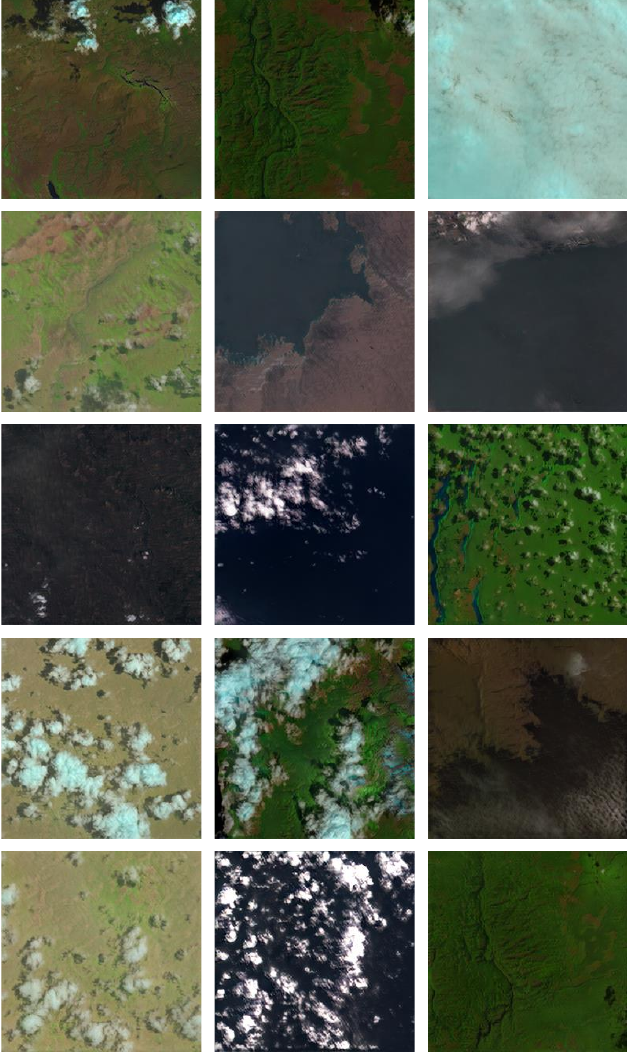}} \hspace{5pt}
    \subfigure[]{\includegraphics[width=0.25\textwidth]{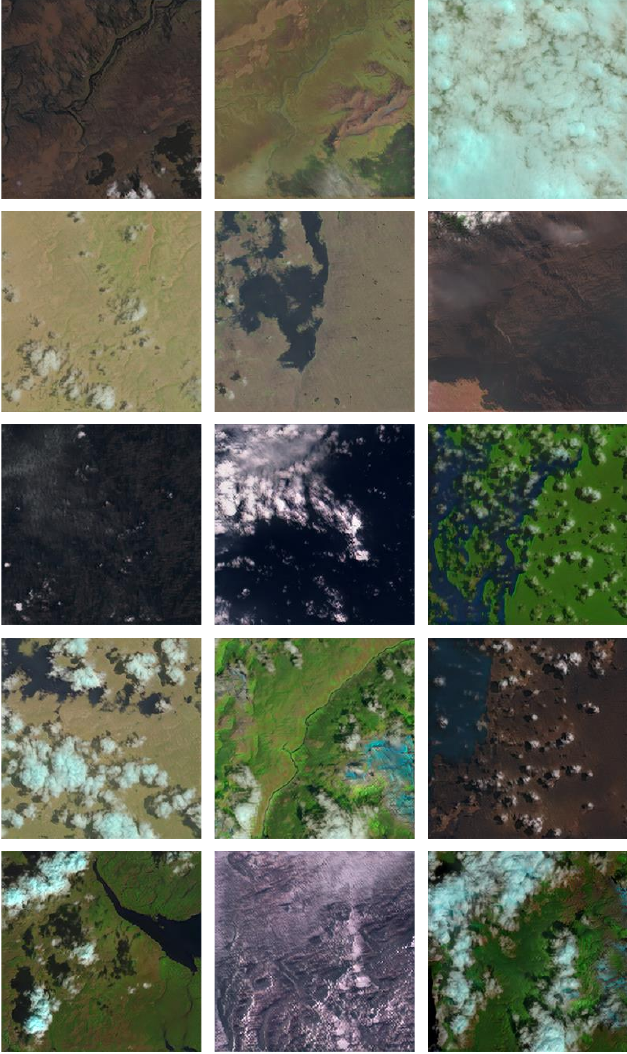}} 
    
\caption{(a) Cloud dataset. (b) Images generated by StyleGAN2+ADA. (c) Images generated by our method.}
\label{fig-cloud}
\end{figure*}

\begin{figure*}[htpb]
    \centering
    \subfigure[]{\includegraphics[width=0.25\textwidth]{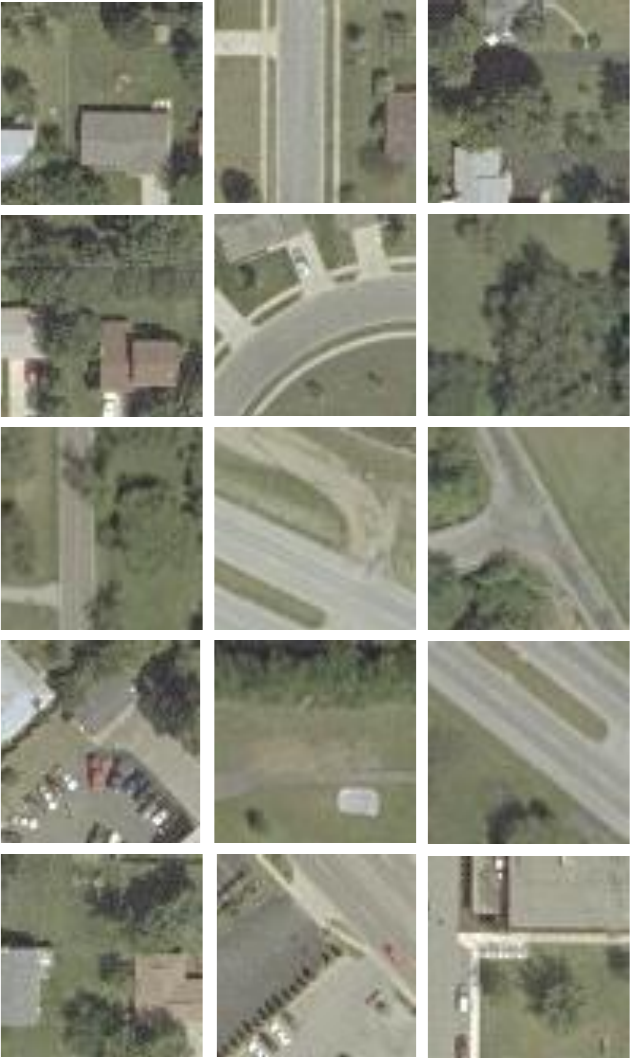}} \hspace{5pt}
    \subfigure[]{\includegraphics[width=0.25\textwidth]{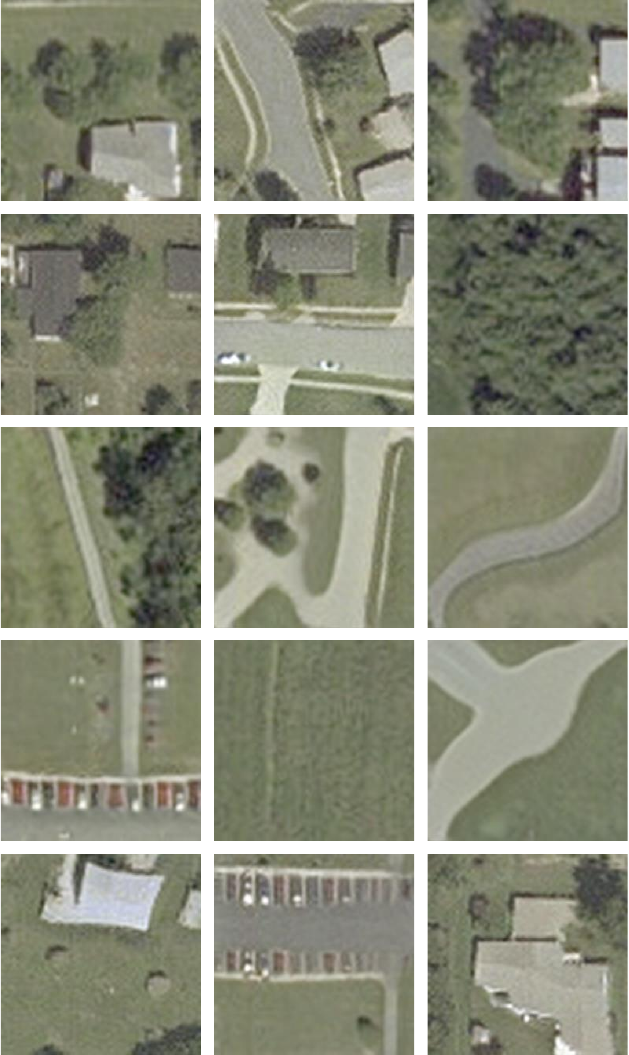}} \hspace{5pt}
    \subfigure[]{\includegraphics[width=0.25\textwidth]{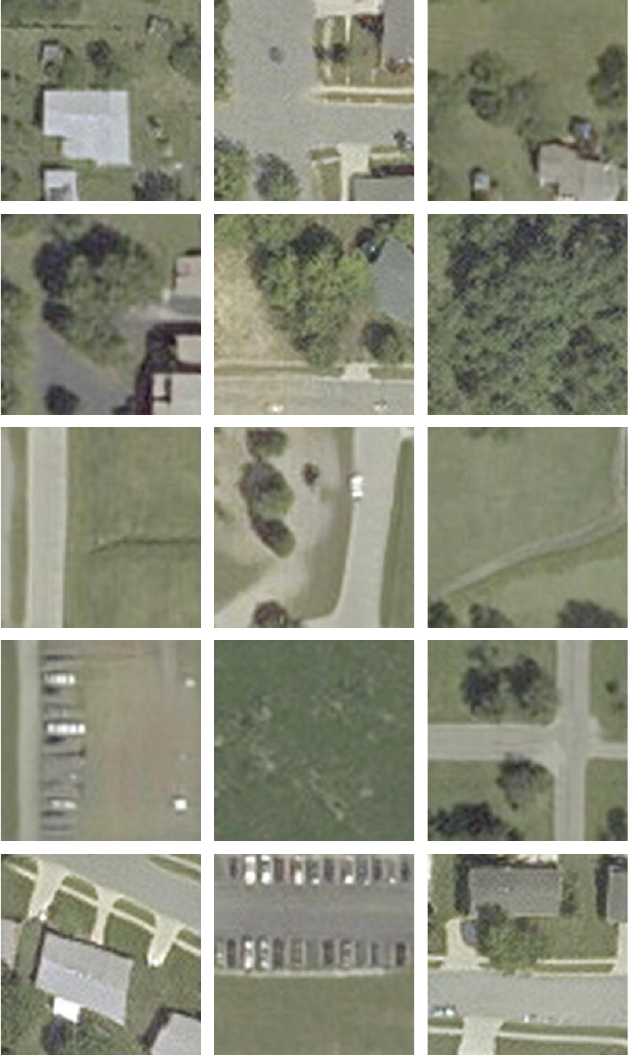}} 
    
\caption{(a) USGS dataset. (b) Images generated by StyleGAN2+ADA. (c) Images generated by our method.}
\label{fig-USGS}
\end{figure*}

\begin{figure*}[htpb]
    \centering
    \subfigure[]{\includegraphics[width=0.25\textwidth]{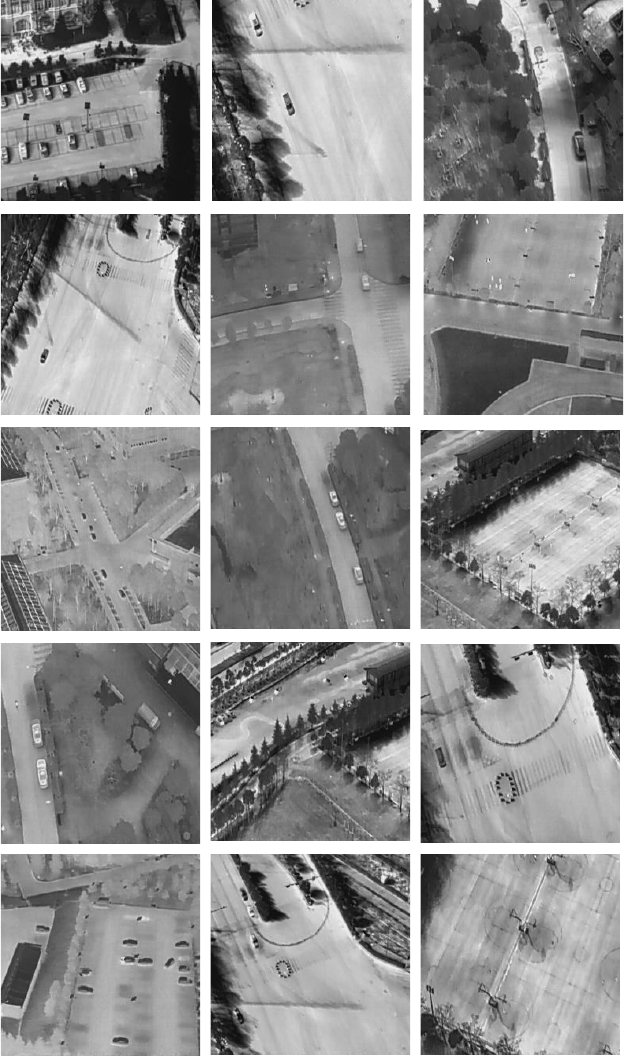}} \hspace{5pt}
    \subfigure[]{\includegraphics[width=0.25\textwidth]{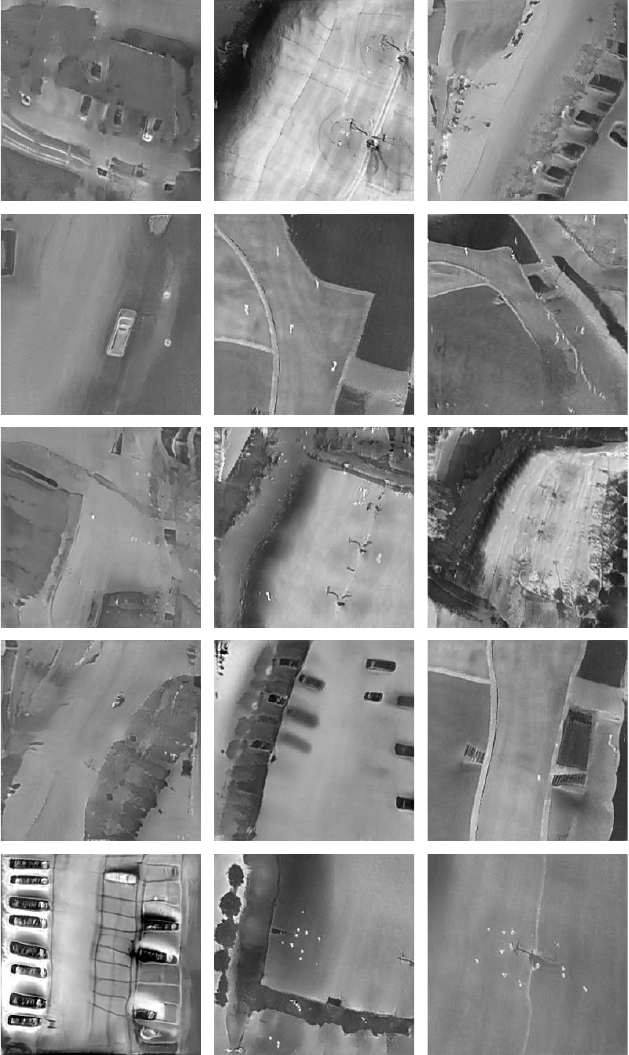}} \hspace{5pt}
    \subfigure[]{\includegraphics[width=0.25\textwidth]{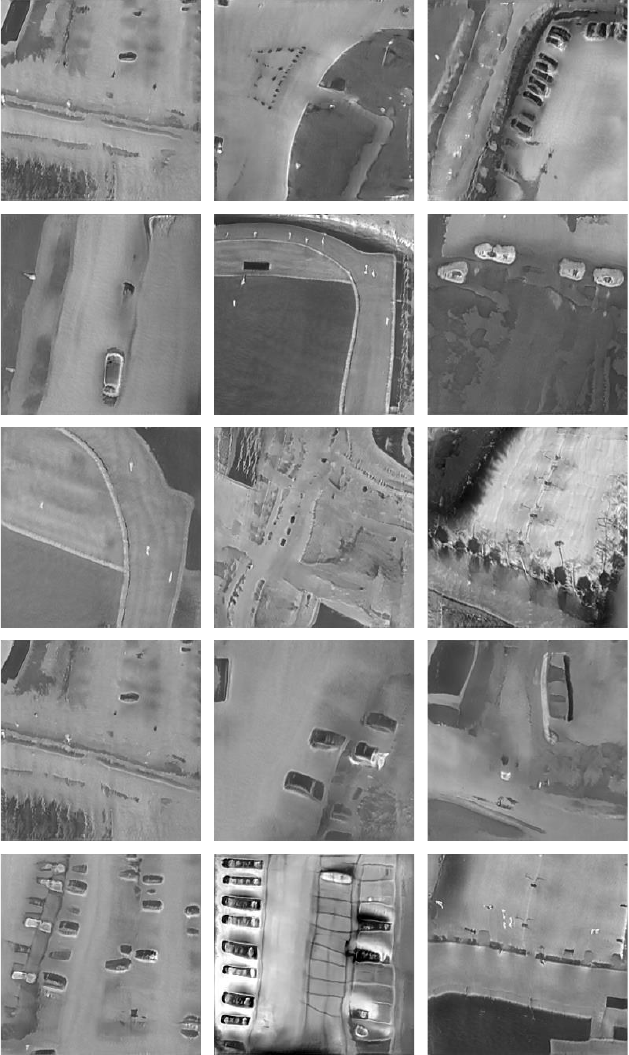}} 
\caption{(a) HIT-UAV dataset. (b) Images generated by StyleGAN2+ADA. (c) Images generated by our method.}
\label{fig-UAV}
\end{figure*}

\begin{figure*}[htpb]
    \centering
    \subfigure[]{\includegraphics[width=0.25\textwidth]{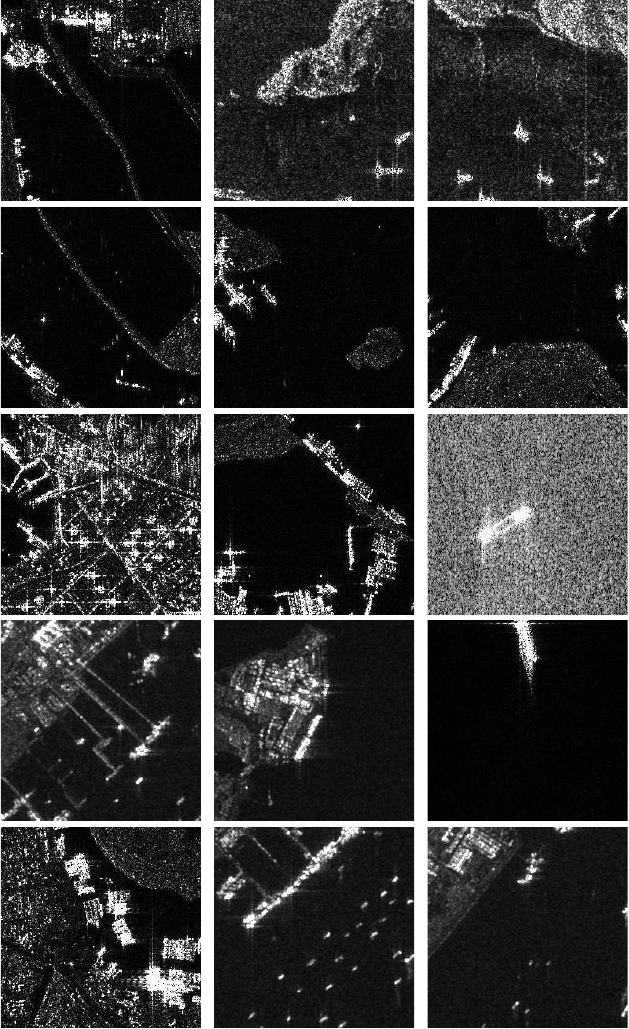}} \hspace{5pt}
    \subfigure[]{\includegraphics[width=0.25\textwidth]{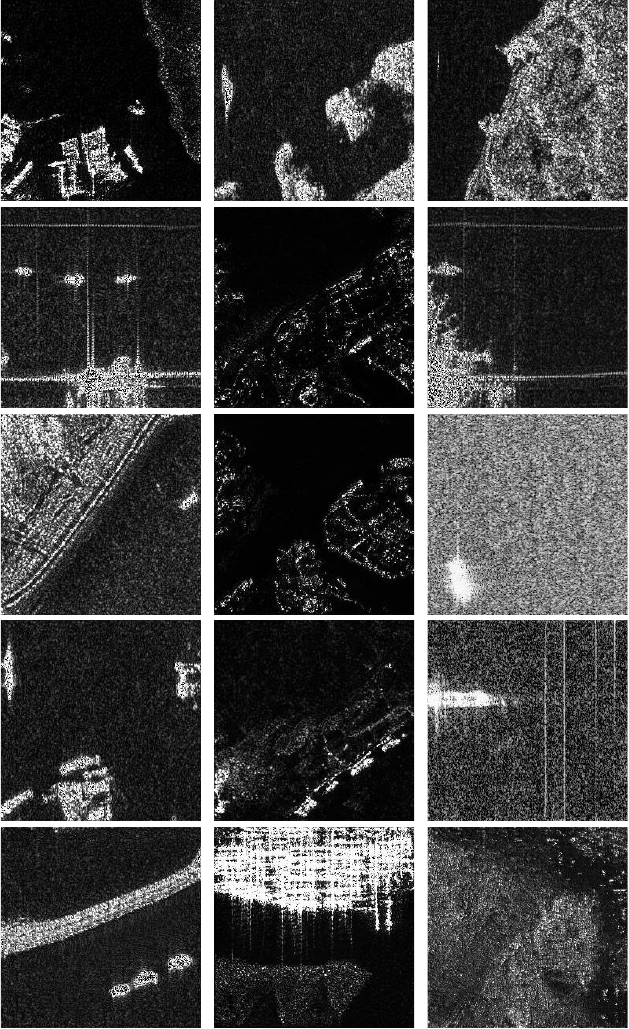}} \hspace{5pt}
    \subfigure[]{\includegraphics[width=0.25\textwidth]{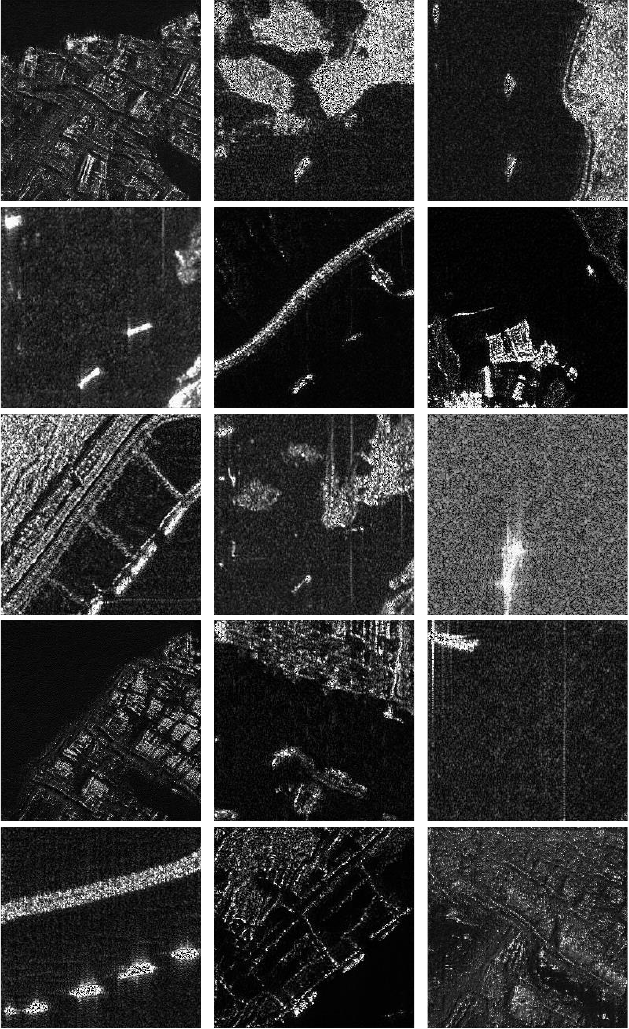}} 
    
\caption{(a) SAR dataset. (b) Images generated by StyleGAN2+ADA. (c) Images generated by our method.}
\label{fig-SAR}
\end{figure*}

\begin{figure*}[htpb]
    \centering
    \subfigure[]{\includegraphics[width=0.25\textwidth]{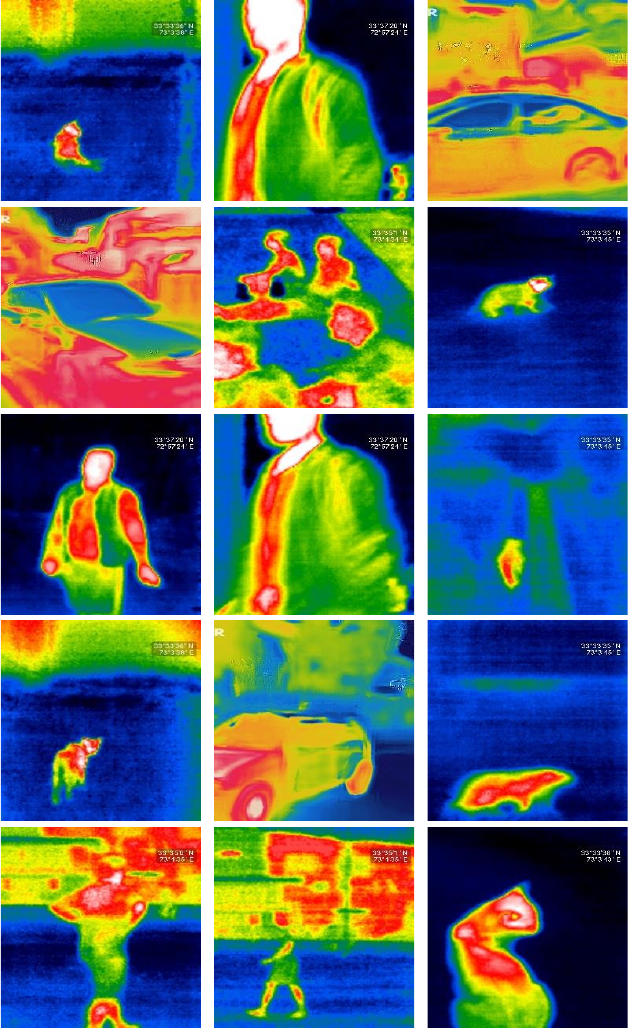}} \hspace{5pt}
    \subfigure[]{\includegraphics[width=0.25\textwidth]{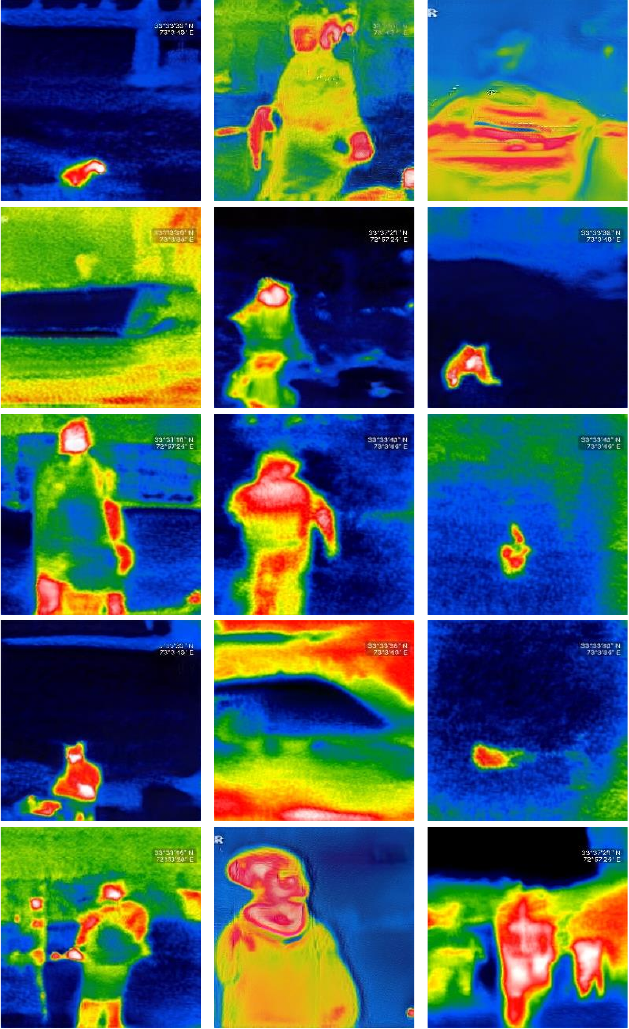}} \hspace{5pt}
    \subfigure[]{\includegraphics[width=0.25\textwidth]{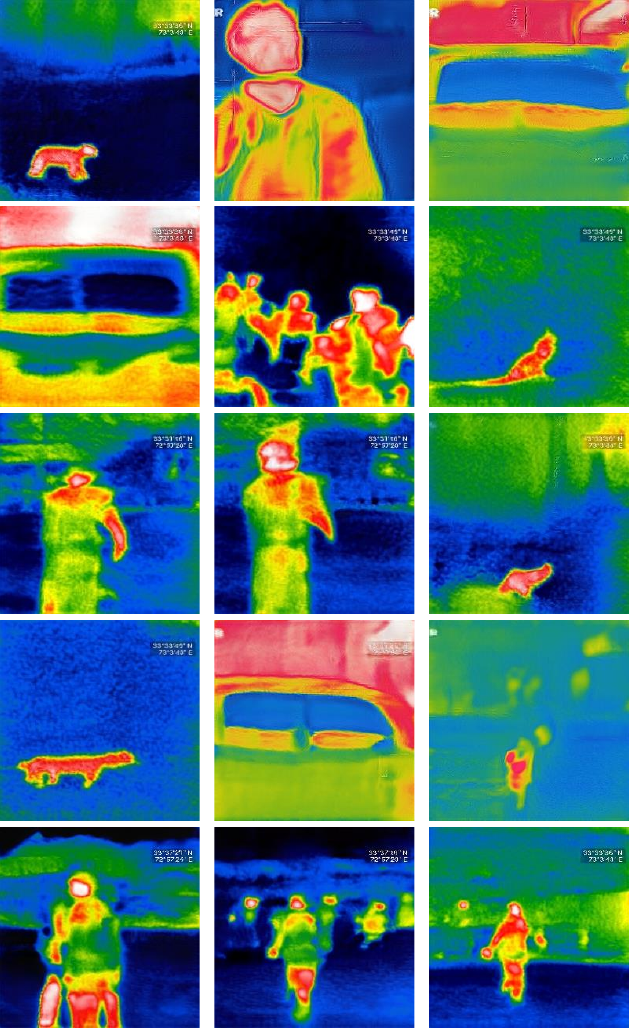}} 
    
\caption{(a) Thermal dataset. (b) Images generated by StyleGAN2+ADA. (c) Images generated by our method.}
\label{fig-thermal}
\end{figure*}

\begin{figure*}[htbp]
\centering
\includegraphics[width=0.8\textwidth]{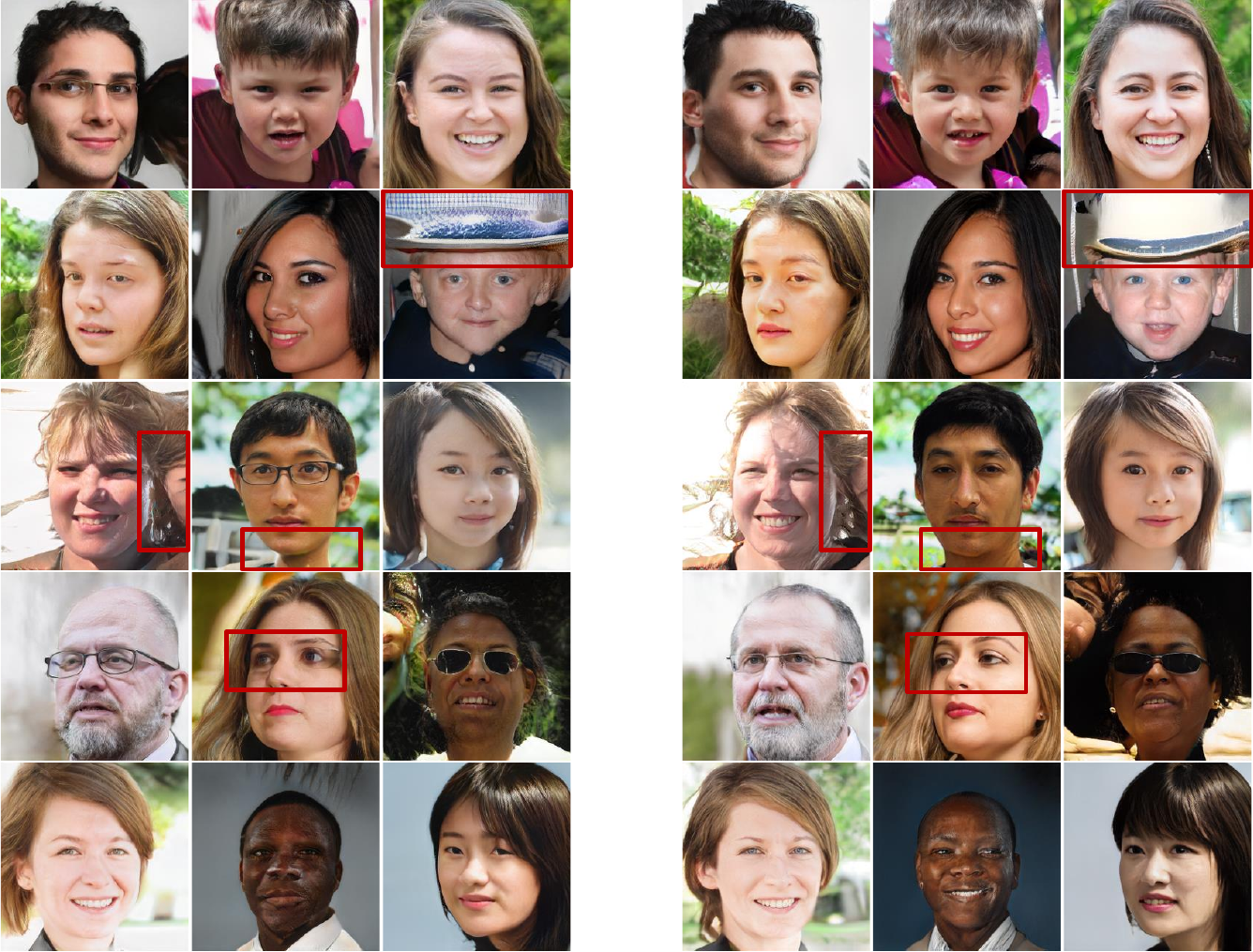}
\caption{The images generated by StyleGAN2+ADA(left), and our method(right) trained on FFHQ256 dataset.}
\label{fig9}
\end{figure*}

\begin{figure*}[htbp]
\centering
\includegraphics[width=0.8\textwidth]{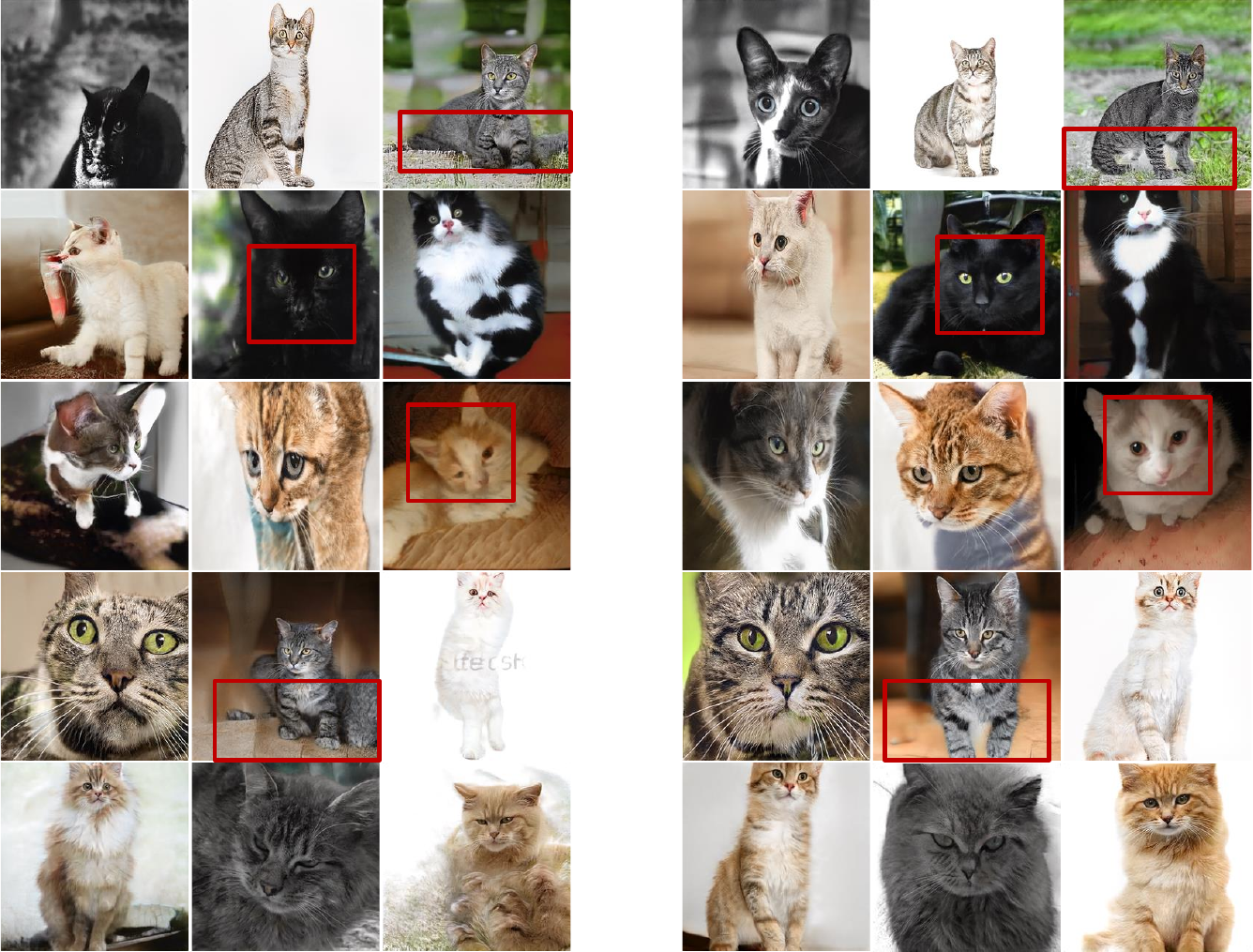}
\caption{The images generated by StyleGAN2+ADA(left), and our method(right) trained on LSUN-cat dataset.}
\label{fig10}
\end{figure*}

The generated images on NWPU, PN and AID datasets are visulized in the Fig.\ref{fig6}, Fig.\ref{fig7} and Fig.\ref{fig8}. Overall, the images generated by the BigGAN gets the worst quality. Compared with BigGAN and StyleGAN2, the images generated by our method have great diversity and rich content. The generated images on Cloud, USGS, HIT-UAV, SAR and Thermal datasets are visualized in the Fig.\ref{fig-cloud}, Fig.\ref{fig-USGS}, Fig.\ref{fig-UAV}, Fig.\ref{fig-SAR} and Fig.\ref{fig-thermal}. Our method are effective on RS images of different scales and different modalities. The generated images on FFHQ256(10k) and LSUN-cat(30k) datasets are visualized in the Fig.\ref{fig9} and Fig.\ref{fig10}. Our method still outperforms StyleGAN2 on these datasets.

\end{document}